\pdfoutput=1
\documentclass[11pt]{article}

\usepackage{times}

\usepackage{latexsym}
\usepackage{url}
\usepackage[T1]{fontenc}
\PassOptionsToPackage{dvipsnames}{xcolor}
\PassOptionsToPackage{table}{xcolor}
\usepackage{emnlp2021}
\usepackage{ytableau,tikz,varwidth}
\usepackage{enumitem}
\usepackage{times}
\usepackage{latexsym}
\usepackage{tabularx}
\usepackage{booktabs, makecell, multirow, rotating}
\usepackage{collcell}
\usepackage{siunitx}
\usepackage{tikz,pgfplots}
\usepackage{fancyhdr,graphicx,amsmath}
\usepackage[linesnumbered,ruled,vlined]{algorithm2e}
\usepackage{etoolbox}

\usepackage{emnlp2021}

\usepackage{tcolorbox}
\usepackage{soul}

\makeatletter
\patchcmd{\@algocf@start}% <cmd>
  {-1.5em}% <search>
  {0pt}% <replace>
  {}{}% <success><failure>
\makeatother

\def\lpmln{${\rm LP}^{\rm{MLN}}$}
\def\beq{\begin{equation}}
\def\eeq#1{\label{#1}\end{equation}}
\def\sm{\hbox{\rm SM}}

\newcommand{\rev}[1]{{{#1}}}%\color{blue!70}{#1}}}

\newcommand{\system}{{\sc RuleBERT}\xspace}
\newcommand{\systemsmall}{{\sc RuleBERT$_{20}$}\xspace}

\newcommand{\fontsmall}{\fontsize{10pt}{12pt}\selectfont}

\newcolumntype{P}[3]{
  >{\collectcell{\colorfromval{#1}{#2}{#3}}} l <{\endcollectcell}}
  
\newcommand*{\colorfromval}[4]{
  \pgfmathparse{#4*50+10}
  \global\let\colorratio\pgfmathresult
  \cellcolor{blue!\colorratio}%
  \tablenum[#3]{#4}}

\newlength\mylen
\newcommand\myinput[1]{%
  \settowidth\mylen{\KwIn{}}%
  \setlength\hangindent{\mylen}%
  \hspace*{\mylen}#1}

\usepackage{adjustbox}
\usepackage{amsmath,amssymb}
\DeclareMathOperator{\EX}{\mathbb{E}}% expected value

\usepackage{microtype}

\title{\system: Teaching Soft Rules to Pre-Trained Language Models}

\author{
Mohammed Saeed$^\dagger$~~~ Naser Ahmadi$^{\dagger}$ ~~~ Preslav Nakov$^\ddagger$ ~~~ Paolo Papotti$^\dagger$\\
       {$^\dagger$Eurecom, France ~~~~~ $^\ddagger$Qatar Computing Research Institute, HBKU, Qatar}\\
       { \{ \textit{FirstName.LastName} \}@eurecom.fr,  pnakov@hbku.edu.qa  }
}

\date{}

\begin{document}
\maketitle
\begin{abstract}
While pre-trained language models (PLMs) are the go-to solution to tackle many natural language processing problems, they are still very limited in their ability to capture and to use common-sense knowledge. In fact, even if information is available in the form of approximate (soft) logical rules, it is not clear how to transfer it to a PLM in order to improve its performance for deductive reasoning tasks. Here, we aim to bridge this gap by teaching PLMs how to reason with soft Horn rules. We introduce a classification task where, given facts and soft rules, the PLM should return a prediction with a probability for a given hypothesis. We release the first dataset for this task, and we propose a revised loss function that enables the PLM to learn how to predict precise probabilities for the task. Our evaluation results show that the resulting fine-tuned models achieve very high performance, even on logical rules that were unseen at training. Moreover, we demonstrate that logical notions expressed by the rules are transferred to the fine-tuned model, yielding state-of-the-art results on external datasets.
\end{abstract}

\section{Introduction}
\label{sec:intro}

\begin{figure}[t]
    \centering
\fontsmall
\noindent\fbox{%
    \parbox{0.95\columnwidth}{%
    \textbf{Input facts}: 
    
    \textit{Mike} is the parent of \textbf{Anne}. \textbf{Anne} lives with Mark. \textbf{Anne} is the child of \underline{Laure}. 
    \textbf{Anne} lives with \textit{Mike}.

    \textbf{Input rules}: 
    
    ($r_1$, .1) Two persons living together are married. 
    
    ($r_2$, .7) Persons with a common child are married. 
    
    ($r_3$, .9) Someone cannot be married to his/her child.
    
    ($r_4$, 1) Every person is the parent of his/her child.
    \vspace{1ex}
    
    \textbf{Test 1}: \underline{Laure} and \textit{Mike} are married. 
    
    \textbf{Answer}: \textit{True} with probability 0.7 [$r_4,r_2$]
    
    \vspace{1ex}
    \textbf{Test 2}: \textbf{Anne} and Mark are married. 
    
    \textbf{Answer}: \textit{False} with probability 0.9 [$r_1$]
    
    \vspace{1ex}
    \textbf{Test 3}: \textbf{Anne} and \textit{Mike} are married. 
    
    \textbf{Answer}: \textit{False} with probability 0.9 [$r_1,r_3,r_4$]
    
    }
}
    \caption{Examples of hypotheses that require reasoning using facts and possibly conflicting soft rules (rule id and confidence shown in brackets).}
    \label{fig:example}
\end{figure}

Pre-trained language models (PLMs) based on transformers~\cite{devlin-etal-2019-bert,roberta2020} are established tools for capturing both linguistic and factual knowledge~\citep{clark2019what,rogers-etal-2020-primer}. However, even the largest models fail on basic reasoning tasks. If we consider common relations between entities, we see that such models are not aware of negation, inversion (e.g., \emph{parent-child}), symmetry (e.g., \emph{spouse}), implication, and composition. 
While these are obvious to a human, they are challenging to learn from text corpora as they go beyond linguistic and factual knowledge~\cite{ribeiro-etal-2020-beyond,kassner-schutze-2020-negated}. 
We claim that such reasoning primitives can be transferred to the PLMs by leveraging logical rules, such as those shown in Figure~\ref{fig:example}. 

While there have been initial attempts to teach reasoning with rules to PLMs~\citep{ruletakers,kassner-etal-2020-pretrained}, such approaches model only a 
subclass of logical rules. 
In fact, current solutions focus on exact rules, i.e., rules that hold in all cases. In reality, most of the rules are approximate, or soft, and thus have a certain confidence of being correct. For example, across the 7,015 logical rules defined on the DBpedia knowledge graph, only 11\% have a confidence above 95\%. In the example, rules $r_1$--$r_3$ are soft, i.e.,~cover knowledge that is not true in all circumstances. 
Consider rule $r_2$, stating that if two persons have a child in common, they are most likely married. As $r_2$ has a confidence of being correct of 0.7, this uncertainty is reflected in the probability of the prediction.

With the above considerations in mind, here we show how to reason over soft logical rules with PLMs. We provide facts and rules expressed in natural language, and we ask the PLM to come up with a logical conclusion for a hypothesis, together with the probability for it being true.

Unlike previous approaches~\citep{ruletakers}, we enable deductive reasoning for a large class of soft rules with binary predicates and an unrestricted number of variables. Our model can even reason over settings with conflicting evidence, as shown in Test 3 in Figure~\ref{fig:example}. In the example, as Anne and Mike live together, they have a 0.1 probability of being married because of soft rule $r_1$. However, we can derive from exact rule $r_4$ that Anne is the child of Mike and therefore they cannot be married, according to soft rule $r_3$. 

To model uncertainty, we pick one flavor of probabilistic logic programming languages, \lpmln, for reasoning with soft rules~\citep{lee2016weighted}. It assigns weights to stable models, similarly to how Markov Logic assigns weights to models. However, our method is independent of the logic programming approach at hand, and different models can be fine-tuned with different programming solutions. Our proposal makes use of synthetic examples that ``teach'' the desired formal behavior through fine-tuning. In particular, we express the uncertainty in the loss function used for fine-tuning by explicitly mimicking the results for the same problem modeled with \lpmln.

Our contributions can be summarized as follows:
\begin{itemize}
    \item \rev{We introduce the problem of teaching soft rules expressed in a synthetic language to PLMs through fine-tuning (modeled as binary classification). 
    \item We create and release the first dataset for this task, which contains 3.2M examples derived from 161 rules describing real common-sense patterns with the target probability for the task obtained from a formal reasoner (Section~\ref{sec:dataset}).}
    
    \item We introduce techniques to predict the correct probability of the reasoning output for the given soft rules and facts. Our solution relies on a revised loss function that effectively models the uncertainty of the rules (Section~\ref{sec:rules}). Our approach handles multi-variable rules and
    nicely extends to examples that require reasoning over multiple input rules.
 
    \item We show that our approach enables fine-tuned models to yield prediction probability very close to that produced by a formal reasoner (Section~\ref{sec:exps}). Our PLM fine-tuned on soft rules, \system, can effectively reason with facts and rules that it has not seen at training, even when fine-tuned with only 20 rules.
    
    \item We demonstrate that our fine-tuning approach effectively transfers knowledge about predicate negation and symmetry to the lower levels of the transformer, which benefits from the logical notions in the rules. In particular, \system achieves new state-of-the-art results on \rev{three external datasets}.

\end{itemize}

The data, the code, and the fine-tuned model are available at \url{http://github.com/MhmdSaiid/RuleBert}.

\section{Related Work}

PLMs have been shown to have some reasoning capabilities~\citep{leap_of_thought}, but fail on basic reasoning tasks~\citep{olympics} and are inconsistent~\citep{Elazar2021MeasuringAI}, especially when it comes to negation~\citep{kassner-schutze-2020-negated}. 

Our work focuses on deductive reasoning. Note that it is different from previous work, e.g.,~on measuring the factual knowledge of PLMs~\cite{petroni-etal-2019-language}, on probing the commonsense capabilities of PLMs at the token or at the sentence level~\cite{cat}, or on testing the reasoning capabilities of PLMs on tasks such as age comparison and taxonomy conjunction~\cite{olympics}. Our work relates to Task \#15 in the bAbI dataset~\cite{babi} and to RuleTakers~\cite{ruletakers}. However, we differ (\emph{i})~by using a larger subclass of first-order logic rules (with more variables and various forms), and (\emph{ii})~by incorporating soft rules. 

Our proposal is different from work on Question Answering (QA) with implicit reasoning based on common-sense knowledge~\cite{clark2019boolq}, as we rely purely on deductive logic from explicitly stated rules.

Our approach also differs from methods that semantically parse natural language into a formal representation on which a formal reasoner can be applied~\citep{liang-2016}, as we directly reason with language. Yet, we are also different from Natural Language Inference (NLI) and textual entailment, which work with text directly, but cannot handle Horn rules \citep{maccartney-manning-2009-extended,2013Dagan}.

Unlike previous work~\citep{yang-2017-nips,hamilton-2018-nips,Minervini_2020}, we do not design a new, ad-hoc module for neural reasoning, but we rely solely on the transformer's capability to emulate algorithms~\cite{wang-etal-2019-learning,Lample2020Deep}.

\section{Background}
\label{sec:background}

\paragraph{Language Models.} We focus on language models pre-trained with bidirectional transformer encoders using masked language modeling~\citep{devlin-etal-2019-bert}. For fine-tuning, we create examples for sequence classification to teach the models how to emulate reasoning given facts and soft rules.

\paragraph{Logical Rules.} 
We rely on existing corpora of declarative \textit{Horn rules} mined from large RDF knowledge bases (KBs)~\citep{galarraga2015fast,OrtonaMP18,AhmadiTDOP20}. An RDF KB is a database representing information with triples (or \textit{facts}) $p(s,o)$, where a \emph{predicate} $p$ connects a \emph{subject} $s$ and an \emph{object} $o$.  An atom in a rule is a predicate connecting two universally quantified variables. A Horn rule (or clause) has the form: $B \rightarrow h(x,y)$, where $h(x, y)$ is a single atom (head or conclusion of the rule) and $B$ (body or premise of the rule) is a conjunction of atoms. Positive rules identify relationships between entities, e.g., $r_1$, $r_2$, $r_4$ in Figure~\ref{fig:example}.
Negative rules identify contradictions, e.g.,~$r_3$ in Figure~\ref{fig:example}. Rules can contain predicates comparing numerical values, such as $<$. 
For example, negative rule
$r_5$: \textit{birthYear(b,d)} $\wedge$ \textit{foundYear(a,c)} $\wedge$ $<$\textit{(c,d)} $\rightarrow$ \textit{negfounder(a,b)} states that any person (variable $b$) with a birth year ($d$) higher than the founding year ($c$) of a company ($a$) cannot be its founder. A fact is derived from a rule if all the variables in the rule body are replaced with constants from facts. For $r_5$, facts ``foundYear(Ford,1903), birthYear(E. Musk,1971), $>$(1971,1903)'' trigger the rule that derives the fact \textit{negFounder}(E. Musk, Ford).

\paragraph{Rule Confidence.} \textit{Exact} rules, such as $r_4$, apply in all cases, without exception. However, most rules are approximate, or soft, as they apply with a certain likelihood. For example, $r_3$ in Figure~\ref{fig:example} is true in most cases, but there are historical exceptions in royal families. Rules are annotated with a measure of this likelihood, either manually or with a computed \textit{confidence}~\citep{galarraga2015fast}. 

\paragraph{Probabilistic Answer Set Programming.}
As we deal with soft rules, we adopt \lpmln~\citep{lee2016weighted} to create the dataset. \lpmln \ is a probabilistic extension of answer set programs (ASP) with the concept of weighted rules from Markov Logic~\cite{chittaBook}. In ASP, search problems are reduced to computing {\em stable models} (answer sets), a set of beliefs described by the program. A weight (or confidence) is assigned to each rule, so that the more rules a stable model satisfies, the larger weight it gets, and the probability of the stable model is computed by normalizing its weight among all stable models. Given a set of soft rules and facts, we measure how much the hypothesis is supported by the stable model.

\section{Dataset}
\label{sec:dataset}

We start by defining the reasoning task. We then discuss example generation methods for three scenarios: single rule as input, multiple (possibly conflicting) rules that require reasoning for the same conclusion, and multiple rules that require a sequence (chain) of reasoning steps. Examples of the data generation procedures are in the Appendix.

\subsection{Reasoning Task}
\label{sec:define_reasoning}

Each example is a triple \textit{(context, hypothesis, confidence)}. \textit{Context} is a combination of rule(s) and generated facts, such as ``\emph{If the first person lives together with the second person, then the first person is the spouse of the second person.}'' and ``\emph{Anne lives with Mike.}'' \textit{Hypothesis} is the statement to be assessed based on the context, e.g., ``\emph{Laure is the spouse of Mike.}'' \textit{Confidence} is the probability that the hypothesis is valid given by the reasoner, e.g.,~0.7. As we generate the examples, we know the confidence for each hypothesis.

\subsection{Single-Rule Dataset Generation}
\label{sec:single_rule_dg}

Given a rule, we generate examples of different hypotheses to expose the model to various contexts. Each example contains the context $c$ and a hypothesis $h$ with its probability of being true as obtained for the ($c,h$) pair from the \lpmln \ reasoner. The intuition is that the examples show the expected behavior of a formal reasoner for every combination of possible facts for a given rule. This process is not about teaching the model specific facts to recall later, but teaching it reasoning patterns.

Unlike previous work~\cite{ruletakers}, our rules allow for multiple variables. This introduces additional complexity as examples must show how to deal with the symmetry of the predicate. For example, \textit{child(Alice,Bob)} and \textit{child(Bob,Alice)} are not equivalent since \textit{child} is not symmetric, while \textit{spouse(Alice,Bob)} and \textit{spouse(Bob,Alice)} are equivalent as \textit{spouse} is symmetric. We assume that metadata about the symmetry and the types is available from the KB for the predicates in the rules. 

Given as input (\emph{i})~a rule $r$, (\emph{ii})~a desired number $n$ of examples, (\emph{iii})~an integer $m$ to indicate the maximum number of facts given as a context, and (\emph{iv})~a pool of values for each type involved in $r$'s predicate \textit{pools}, Algorithm~\ref{alg:dg_alg} outputs a dataset $D$ of generated examples.  
 
\let\oldnl\nl% Store \nl in \oldnl
\newcommand{\nonl}{\renewcommand{\nl}{\let\nl\oldnl}}% Remove line number for one line

\SetInd{0.3em}{0.18em}
\begin{algorithm}[tbh]
\caption{Generate Synthetic Data}
\label{alg:dg_alg}
\SetKwProg{Fn}{Function}{:}{}
\SetKwFunction{FALTER}{Alter}
\SetKwFunction{FGenFacts}{GenFacts}

\small
\KwIn{rule $r$ \tcp*{\footnotesize\ttfamily{child(a,b)$\rightarrow$parent(b,a)}}} 
\nonl \myinput{$n$ \tcp*{\# of examples}}
\nonl \myinput{$m$  \tcp*{max \# of facts}}
\nonl \myinput{$pools$ \tcp*{pools of names}}

\KwOut{Generated Dataset $D$}

$D=\{\}$, $i = 1$ \tcp*{\footnotesize\ttfamily{initialize}}
\While{$i \leq ceiling(n/8)$}{
    $F = GenFacts(r,m,pools)$ \tcp*{\footnotesize\ttfamily{\textit{child(Eve,Bob),parent(Amy,Sam)}}}
    $O=LPMLN(r,F)$ \tcp*{\footnotesize\ttfamily{reasoner output}}
    $h_{1} = f \in F$ \tcp*{\footnotesize\ttfamily{\textit{child(Eve,Bob)}}}
    $h_{2}=Alter(f)$ \tcp*{\footnotesize\ttfamily{\textit{negchild(Eve,Bob)}}}
    $h_{3}= r(F)$ \tcp*{\footnotesize\ttfamily{\textit{parent(Bob,Eve)}}}
    $h_{4}= Alter(r(F))$ \tcp*{\footnotesize\ttfamily{\textit{parent(Eve,Bob)}}}
    $h_{5} =$ pos.$f_l \notin F$ \tcp*{\footnotesize\ttfamily{\textit{child(Joe,Garry)}}}
    $h_{6} = \neg h_{5}$ \tcp*{\footnotesize\ttfamily{\textit{negchild(Joe,Garry)}}}
    $h_{7} = f_r \notin O$ \tcp*{\footnotesize\ttfamily{parent(Alice,Joe)}}
    $h_{8} = \neg h_{7}$ \tcp*{\textit{negparent(Alice,Joe)}}
    $D.add(h_{1-8})$\;
    $i \leftarrow i + 1$\;
    }
\Fn{\FGenFacts{r,m,pools}}{ 
$F=GetRandomFacts(r,pools,m)$\;
$F.add(GetRuleFacts(r,pools))$\;
\textbf{return} F
}
\Fn{\FALTER{$p(s,o)$}}{
    \lIf{p is symmetric}
        {\textbf{return}  $ \neg p(s,o)$
        }
    {\lIf{random()$>$0.5}  
        {\textbf{return}  $\neg p(s,o)$ 
        }    
        \lElse {\textbf{return} $p(o,s)$
        }
        }
}
\end{algorithm}
  
We start at line 3 by generating facts, such as \textit{child(Eve,Bob)}, using the function \textit{GenFacts} (lines 15--18), which takes as input a rule $r$, the maximum number of facts $m$ to generate, and the \textit{pools}. A random integer less than $m$ sets the number of facts in the current context. The generated facts $F$ have predicates from the body of $r$, their polarity (true or negated atom) is assigned randomly, and variables are instantiated with values sampled from the pool (line 16). Facts are created randomly, as we are not interested in teaching the model specific facts to recall later, but instead we want to teach it how to reason with different combinations of rules and facts. We then ensure that the rule is triggered in every context, eventually adding more facts to $F$ using the function \textit{GetRuleFacts} in line 17. After obtaining $F$, we feed rule $r$ along with facts $F$ to the \lpmln \ reasoner, and we obtain a set $O$ containing all satisfied facts and rule conclusions (line 4).

We generate different hypotheses, where each one leads to an example in dataset $D$. For each context, we add an example with different facts with respect to the given rule according to three dimensions. A fact can be (\emph{i})~for a predicate in the premise or in the conclusion of a rule, could be (\emph{ii})~satisfied or unsatisfied given the rule, and could have (\emph{iii})~positive or negative polarity. This makes eight different possibilities, thus leading to the generation of eight different hypotheses (one for each context). 

The first hypothesis $h_{1}$ is obtained by sampling a fact from the set $F$ (line $5$). We then produce the counter hypothesis $h_{2}$ by altering the fact (line 6) using the function \textit{Alter} (lines 19-22). Given a hypothesis $p(s,o)$ (line 19), we return its negated form if $p$ is symmetric (line 20). Otherwise, if $p$ is not symmetric, we produce a counter hypothesis either by negation (line 21), or by switching the subject and the object in the triple as the predicate is not symmetric (line 22). We rely on a dictionary to check whether a predicate is symmetric or not.

We then produce hypothesis $h_{3}$ (line 7), which is the outcome of triggering rule $r$ with the facts added in line 17. The counter hypothesis $h_{4}$ is generated by altering $h_{3}$ (line 8). Moreover, we generate hypothesis $h_{5}$ by considering any unsatisfied positive fact outside $F$. Following a closed-world assumption (CWA), we assume that positive triples are false if they cannot be proven, meaning that their negation is true. We sample a fact $f_{l}$ from the set of all possible positive facts that do not have the same predicate of the rule head (line 9). Thus, $h_{5}$ will never be in the output $O$ of the reasoner, as it cannot be derived. We then produce $h_{6}$ by negating $h_{5}$ in line 10. We further derive $h_{7}$ by sampling a fact $f_{r}$ that has the same predicate as that of the rule head, but does not belong to the output of the reasoner $O$ (line 11). For a positive (negative) rule, such a fact is labelled as False (True). $h_{7}$ is then negated to get the counter hypothesis $h_{8}$ (line 12). All generated hypotheses are added to $D$ (line 13), and the process repeats until we obtain $n$ examples.

Finally, we automatically convert the examples to natural language using predefined templates. A basic template for atom predicate $p$ (type $t_1$, type $t_2$) is ``If the 1$^{st}$ $t_1$ is $p$ of the 2$^{nd}$ $t_2$.''
(``If the first person is spouse of $\dots$''). For the single-rule scenario, we release a dataset for 161 rules with a total of 3.2M examples and a 80\%:10\%:10\% split for training, validation, and testing.

\subsection{Rules with Overlapping Conclusion}
\label{sec:dg_union}

When multiple rules are in the context, there could be facts that trigger more than one rule for a given hypothesis. The triggered rules might be all of the same polarity (positive or negative), eventually accumulating their confidence, or could be a mix of positive and negative rules that oppose each other. 
%In either case, the model needs to learn the interaction between the soft rules. 
While the data generation procedure in Section~\ref{sec:single_rule_dg} can be extended to handle multiple rules, this raises an efficiency problem.
Given a set of $R$ rules, it would generate $8^{|R|}$ examples for each \textit{(facts,rule)} pair in order to cover all rule combinations. This is very expensive, e.g.,~for five rules, it would generate $8^{5}= 32,768$ examples for a single context. 

Given this challenge, we follow a different approach.
%when generating data for multiple rules having the same predicate 
%(or its negation) 
%in their conclusion. 
We first generate data for each rule individually using Algorithm~\ref{alg:dg_alg}. %The resulting \rev{examples contain} %datasets teach the model how to deal with 
%satisfied and unsatisfied facts, and rule conclusions for each rule. 
We then generate more examples only for combinations of two or more rules having \textit{all their} \textit{\rev{rule} conclusions} %(and their negation/switching) 
as hypotheses. 
%These examples \rev{show} what to predict when two or more rules are triggered. 
%An illustration of such examples is shown in the Appendix in section~\ref{sec:rule_overlap_example}.
%, while it learns how to deal with  the other cases from the single-rule datasets. 
For every input context, we produce rule-conclusion hypotheses (positive and negative) while varying 
%each time 
the rules being fired.
Thus, %for $r$ rules, 
we generate $2 * \sum_{x=2}^{|R|} {|R| \choose x}$ examples with at least two rules triggered. Adding the single-rule data, we generate $ 8*|R| +2 * \sum_{x=2}^{|R|} {|R| \choose x}$ for every \textit{(facts,rules)} pair, which is considerably smaller than $8^{r}$ for $|R|\geq 2$, according to the binomial theorem. For example, for $|R|$=5, we generate 92 examples per context. %, which is considerably less expensive. 
\rev{For the overlapping rules scenario, we release a dataset for 5 rules with a total of 300K examples, and a 70\%:10\%:20\% split for training, validation, and testing.}

\subsection{Chaining of Rule Executions}
\label{subsec:dg_chain}

For certain hypotheses, an answer may be obtained by executing rules in a sequence, i.e.,~one on the result of the other, or in a \textit{chain}. To be able to evaluate a model in this scenario, we generate hypotheses that can be tested only by chaining a number of rules (an example is shown in Appendix~\ref{sec:chain_example}). Given a pool of rules over different relations and a depth $D$, we sample a chain of rules with length $D$. We then generate hypotheses that would require a depth varying between 0 and $D$. We generate a rule-conclusion hypothesis ($h_{3}$) and its alteration ($h_{4}$) for each depth $d \leq D$. A depth of 0 means that the hypothesis can be verified using the facts alone without triggering any rule. We also generate counter-hypotheses by altering the hypotheses at a given depth, and we further include hypotheses that are unsatisfied given the input.

For the chaining rules scenario, we start with a pool of 64 soft rules, and we generate hypotheses that would need at most five chained rules to verify them. The dataset for  $d \leq 5$ contains a total of 70K examples, and a 70\%:10\%:20\% split for training, validation, and testing.

\section{Teaching PLMs to Reason}
\label{sec:rules}

In this section, we explain how we teach a PLM to reason with one or more soft rules. Note that uncertainty stems from the rule confidence. One approach to teach how to estimate the probability of a prediction is to treat each confidence value (or bucket of confidence values) as a class and to model the problem as a $k$-way classification instance (or regression), but this is intractable when multiple rules are considered. Instead, we keep the problem as a two-class one by altering how the information is propagated in the model to incorporate uncertainty from the rule confidence.

Let $D = \{(x_{i},y_{i})\}^{m}_{i=1}$ be our generated dataset, where $x_{i}$ is one example of the form \textit{(context,hypothesis,confidence)} and $y_{i}$ is a label indicating whether the hypothesis is validated or not by the context (facts and rules in English), and $m$ is the size of the training set. A classifier $f$ is a function that maps the input to one of the labels in the label space. Let $h(x,y)$ be a classification loss function. The empirical risk of the classifier $f$ is

{\small
\begin{align*}
R_{h}(f)= \EX_{D}(h(x,y))=-\frac{1}{m} \sum_{i=1}^{m} h(x_{i},y_{i})
\end{align*}
}
We want to introduce uncertainty in our loss function, using the weights computed by the \lpmln \ solver as a proxy to represent the probability of predicting the hypothesis as being true. To do so, we apply a revised empirical risk:

{\small
\begin{flalign*}
  &R'_{h}(f)= \EX_{D}(h(x,y))= &\\
 & -\frac{1}{m} \sum_{i=1}^{m} (w(x_{i}) * h(x_{i},1)+
  (1-w(x_{i})) * h(x_{i},0))&
\end{flalign*}
}
\noindent where $w(x_{i})$ is the probability of $x_{i}$ being True.

We now state that each example is considered as a combination both of a weighted positive example with a weight $w(x_{i})$ provided by the \lpmln \ solver and a weighted negative example with a weight $1 - w(x_{i})$. 

When trained to minimize this risk, the model learns to assign the weights to each output class, thus predicting the confidence for the true class when given the satisfied rule head as a hypothesis.

\section{Experiments}
\label{sec:exps}

We first describe the experimental setup (Section~\ref{sec:exp_setup}). We then evaluate the model on single (Section ~\ref{sec:single_exp}) and on multiple rules (Sections~\ref{sec:exp_union} and \ref{sec:exp_chain}). We show that a PLM fine-tuned on soft rules, namely \system, makes accurate predictions for unseen rules (Section~\ref{sec:new_rules_exp}), and it is more consistent than existing models on three external datasets (Section~\ref{sec:neg_lama}). We report the values of the hyper-parameters, as well as the results for some ablation experiments in the Appendix. The datasets for all experiments are summarized in Table~\ref{tab:dataset_distributions}.

\subsection{Experimental Setup}
\label{sec:exp_setup}

\paragraph{Rules.} We use a corpus of 161 soft rules mined from DBpedia. We chose a pool of distinct rules with varying number of variables, number of predicates, rule conclusions, and confidences.

\paragraph{Reasoner.} We use the official implementation\footnote{\url{http://github.com/azreasoners/lpmln}} of the \lpmln \ reasoner. We set the reasoner to compute the exact probabilities for the triples.

\paragraph{PLM.} We use the HuggingFace pre-trained \textit{RoBERTa}$_{{\footnotesize LARGE}}$~\citep{roberta2020} model as our base model, as it is trained on more data compared to \textit{BERT}~\citep{devlin-etal-2019-bert}, and is better at learning positional embeddings~\citep{wang-chen-2020-position}. We fine-tune the PLM\footnote{The {prompt} is \textit{<s>context</s></s>hypothesis</s>}.} with the weighted binary cross-entropy (wBCE) loss from Section~\ref{sec:rules}. More details can be found in Appendix~\ref{sec:exps_details}.

\begin{table}[t]
    \centering
    \small
    \begin{tabular}{@{}l@{ }r@{ }@{ }r@{ }@{ }r@{ }@{ }r@{}}
        \toprule
         \textbf{Dataset}& \textbf{Total} & \textbf{Train}& \textbf{Dev}&\textbf{Test}\\
         \midrule
         Single Rule (Section~\ref{sec:single_exp})&20K & 16K &2K&2K\\
         Overlap (Section~\ref{sec:exp_union}) &300K &210K&30K&60K\\
         Chaining (Depth=5) (Section~\ref{sec:exp_chain})&70K &56K&4.6K&9.4K\\
         \system (Section~\ref{sec:neg_lama}) & 3.2M & 2.56M & .32M& .32M\\
         \bottomrule
    \end{tabular}
    \caption{Datasets for the experiments and their splits.}
    \label{tab:dataset_distributions}
\end{table}

\begin{table*}[t]
\centering
\small
\begin{tabular}{l|c|ccc|ccc}
\toprule
 & & \multicolumn{3}{c|}{\textit{RoBERTa-wBCE}}  & \multicolumn{3}{c}{\textit{RoBERTa}}         \\
\textbf{Rule} & {\footnotesize Conf.} & Acc. & \multicolumn{2}{c|}{CA@k} & Acc. & \multicolumn{2}{c}{CA@k} \\
&            &      & .10  & .01   &      & .10   & .01   \\ 
\midrule

birthYear(a,c) $\wedge$ deathYear(b,d) $\wedge$ $>$(c,d)$\rightarrow$negspouse(a,b) &  
.990 & .995   &  .993      &  .993  &   .970    
&  .490   & .486                      \\

birthYear(b,d) $\wedge$ foundYear(a,c) $\wedge$ $<$(c,d)$\rightarrow$negfounder(a,b) & 
.990 &  .928     &    .927    &  .927      &    .908   
&  .486    &    .456     \\ 

spouse(c,a) $\wedge$ parent(b,c) $\rightarrow$ negspouse(a,b) &         
.923 &  .974     &    .963    &  .747     &     .875
& .491    & .279    \\ 

relative(a,c) $\wedge$ spouse(b,c) $\wedge$ child(b,a) $\rightarrow$ relative(a,b) &            
.860 & .922  & .844   &   .801     &    .866
& .342   & .146    \\ 

parent(c,a) $\wedge$ child(b,c) $\rightarrow$ spouse(a,b)   &
.825 &    .944  &    .828  &    .444    &   .842
&  .342  & .146         \\ 

publisher(c,b) $\wedge$ subsequentWork(c,a) $\rightarrow$ publisher(a,b) &            
.721 &     .909  &    .834    &  .765     & .905
& .358     & .219\\ 

successor(b,a) $\rightarrow$ negspouse(a,b) &            
.718 &    .972  &    .896    &  .693  & .949
& .369    & .313 \\ 

child(c,b) $\wedge$ relative(c,a) $\rightarrow$ negchild(a,b) &            
.644 & .935   &  .880    &  .693    & .905
&   .310     &  .303      \\ 

child(c,b) $\wedge$ spouse(a,c) $\rightarrow$ negrelative(a,b)     &            
.562 &  .920  &   .907  &    .608    &  .915
&     .255    &   .250     \\ 

relation(a,b) $\rightarrow$ negchild(a,b)     &            
.549  & .904  &   .886    &  .737     & .902
& .371  & .366      \\ 

child(c,b) $\wedge$ spouse(c,a) $\rightarrow$ child(a,b)     &            
.492 &  .901     &  .827   &   .422     &   .658
&  .223  & .107   \\ 

knownFor(b,a) $\rightarrow$ founder(a,b)     &            
.387 &  .882    & .601      &    .477    &  .839
&     .372    &  .215      \\ 

founder(c,b) $\wedge$ publisher(c,a) $\rightarrow$ negfounder(a,b)     &            
.246 &  .886     &  .795    &  .665      &  .802
&     .311    &  .297      \\ 

publisher(a,c) $\wedge$ parentCompany(b,c) $\rightarrow$ negpublisher(a,b)     &            
.235 &  .812    &  .748      &    .643    &  .811
&   .313      &     .271   \\ 

successor(c,a) $\wedge$ spouse(c,d) $\wedge$ successor(d,b)$\rightarrow$spouse(a,b)     &
.221 &  .927     &  .738     &   .628    &   .761
&  .248    &  .215       \\ 

relative(a,c) $\wedge$ parent(c,b) $\rightarrow$ child(a,b)      &            
.135  &  .841     &  .704   &   .552    &   .727
&  .227       &    .182   \\
\bottomrule
\end{tabular}
\caption{\label{tab:singles}Evaluation results for single-rule models.}
\end{table*}

\paragraph{Evaluations Measures.}
For the examples in the test set, we use accuracy (Acc) and F1 score (F1) for balanced and unbalanced settings, respectively.
As these measures do not take into account the uncertainty of the prediction \textit{probability}, we further introduce Confidence Accuracy@k (CA@k), which measures the proportion of examples whose absolute error between the predicted and the actual probabilities is less than a threshold $k$:
\begin{equation*}
   CA@k=\frac{\#\{x_{i}, |w_{i}-\hat w_{i}| < k \}}{\#\{x_{i}\}}
\end{equation*}
where $x_{i}$ is the i$^{th}$ example of dataset, $w_{i}$ is the actual confidence of the associated hypothesis given by the \lpmln \ reasoner, $\hat w_{i}$ is the predicted confidence by the model, and $k$ is a chosen threshold. 

The measure can be seen as the ordinary accuracy measure, but true positives and negatives are counted only if the condition is satisfied, where lower values for $k$ indicate stricter evaluation.

\subsection{Single Soft Rule} 
\label{sec:single_exp}

We fine-tune \rev{16} models for 16 different positive and negative rules (one model per rule) using 16k training samples per rule. We compare the accuracy of each model (\emph{i})~without teaching uncertainty using binary cross-entropy (\rev{RoBERTa}), and (\emph{ii})~with teaching soft rules using wBCE.

\paragraph{Results.} Every row in Table~\ref{tab:singles} shows a rule with its confidence, followed by accuracy and CA@$k$ (for $k=0.1$ and $k=0.01$) for both loss functions. We see that models fine-tuned using \textit{RoBERTa-wBCE} perform better on CA@k. In terms of \textit{accuracy}, both models perform well, with \textit{RoBERTa-wBCE} performing better for all rules. Interestingly, the best performing rules are two rules that involve comparison of numerical values (birth years against death and founding years), which suggests that our method can handle comparison predicates.

\begin{table}[th]
\small
\begin{tabular}{cccccc}
\toprule
\bf{Test} &  \bf{Size} & \bf{F1} & \bf{{\footnotesize CA@.15}} &  \bf{{\footnotesize CA@.1}} &  \bf{{\footnotesize CA@.05}}  \\
\midrule
$r_1$       &   1.6k &     .990 &                 .987 &        .986 &                 .954 \\
$r_2$       &   1.6k &     .999 &                 .997 &        .996 &                 .946 \\
$r_3$       &   1.6k &     .995 &                 .994 &        .994 &                 .992 \\
$r_4$       &   1.6k &     .990 &                 .989 &        .988 &                 .935 \\
$r_5$       &   1.6k &     1 &                 .999 &                .998 &                 .979 \\
U=2 &  20k &     .985 &                 .997 &                .993 &                 .968 \\
U=3 &  20k &     .925 &                 1 &                .998 &                 .949 \\
U=4 &  10k &     .956 &                 1 &                1 &                 .988 \\
U=5 &   2k &     1 &                 1 &                1 &                 .980 \\
\bottomrule

\end{tabular}
\caption{Results for a model trained on five rules sharing the same predicate, and tested on multiple test sets.}
\label{tab:union_exp_res}
\end{table}

\subsection{Rules Overlapping on Conclusion}
\label{sec:exp_union}

The dataset contains five soft rules with \textit{spouse} or \textit{negspouse} in the rule conclusion, and a confidence between 0.30 and 0.87  (shown in Figure~\ref{fig:union_exp_rules}). We train a model on the dataset and test it (\emph{i})~on a test set for each of the five rules separately, (\emph{ii})~on test sets with $U$ triggered rules, where $U \in \{2,3,4,5\}$.

\begin{figure}[h]
    \centering
\small
\noindent\fbox{%
    \parbox{0.97\columnwidth}{
    ($r_1$, .87) child(a,c) $\wedge$ parent(c,b) $\rightarrow$  spouse(A,B) 
    
    ($r_2$, .64) child(a,b) $\rightarrow$ negspouse(a,b) 
    
    ($r_3$, .3)  relative(a,b) $\rightarrow$ spouse(a,b) 
    
    ($r_4$, ,78) child(a,c) $\wedge$ child(b,c) $\rightarrow$ spouse(a,b) 
    
    ($r_5$, .67) predecessor(a,b) $\rightarrow$ negspouse(a,b) 
    }
}
    \caption{The five overlapping soft rules.}
    \label{fig:union_exp_rules}
\end{figure}

\paragraph{Results.}  Table~\ref{tab:union_exp_res} shows that the model achieves high scores both on the single test sets \rev{(top five rows)} and on the sets with interacting rules. The test sets with $U=2$ and $U=3$ are most challenging, as they contain ${5 \choose 2}=10 $ and ${5 \choose 3}=10 $ combinations of rules, respectively, while the one with $U=5$ has only one possible rule combination. The high scores indicate that PLMs can actually learn the interaction between multiple soft rules.

\subsection{Rule Chaining}
\label{sec:exp_chain}

Here, we assess models fine-tuned on various chaining depths. We construct six datasets for this scenario with increasing depths ($D=0$, $D\leq 1$, $D\leq 2$, $D\leq 3$, $D\leq 4$, $D\leq 5$), i.e.,~dataset $D\leq x$ contains hypotheses that need at most $x$ chained rules. We thus train six models (one per dataset), and we test them (\emph{i})~on their own test dataset (Test), (\emph{ii})~on the test set with $D \leq 5$ that contains all examples up to depth 5 (All), and (\emph{iii})~on test sets with a chaining of depth $x$ (Dep$x$).

\begin{table}[th]
\hspace{1ex}
\footnotesize
\begin{tikzpicture}[inner sep=0in,outer sep=0in]
\node (n) {\begin{varwidth}{12cm}{
\resizebox{0.6\columnwidth}{!}{%
\sisetup{add-integer-zero=false}
\begin{tabular}{l
                P{0}{1}{scientific-notation = false,table-format = 1.2e-0}
                P{0}{1}{scientific-notation = false,table-format = 1.2e-0}
                P{0}{1}{scientific-notation = false,table-format = 1.2e-0}
                P{0}{1}{scientific-notation = false,table-format = 1.2e-0}
                P{0}{1}{scientific-notation = false,table-format = 1.2e-0}
                P{0}{1}{scientific-notation = false,table-format = 1.2e-0}}
\toprule
\bf{Data} &  \multicolumn{1}{l}{\bf Mod0} &  \multicolumn{1}{l}{\bf Mod1} &  \multicolumn{1}{l}{\bf Mod2} &  \multicolumn{1}{l}{\bf Mod3} &  \multicolumn{1}{l}{\bf Mod4} &  \multicolumn{1}{l}{\bf Mod5} \\
\midrule
Test  &  .996 &  .926 &  .883 &  .852 &  .856 &  .831 \\
\hline
All &  .589 &  .743 &  .772 &  .811 &  .831 &  .831 \\
\hline
 Dep0    &  .993 &  .974 &  .973 &  .982 &  .978 &  .973 \\
Dep1    &  .264 &  .860 &  .884 &  .887 &  .889 &  .889 \\ 
Dep2    &  .396 &  .655 &  .730 &  .751 &  .750 &  .720 \\
Dep3    &  .438 &  .581 &  .636 &  .684 &  .690 &  .656 \\
Dep4    &  .538 &  .468 &  .547 &  .626 &  .666 &  .627 \\
Dep5    &  .552 &  .356 &  .496 &  .703 &  .785 &  .744 \\ 
\hline
\end{tabular}
}}
\end{varwidth}};
 \draw [very thick,red] (-3.5,0) -- (-1.6,0) -- (-1.6,-0.3) -- (-0.6,-0.3)
 -- (-0.6,-0.6) -- (0.45,-0.6) -- (0.45,-0.9) -- (1.45,-0.9)--(1.45,-1.2)--(2.5,-1.2)
 --(2.5,-1.5) -- (-3.5,-1.5) -- (-3.5,0);

\end{tikzpicture}
\caption{F1 scores for models trained on varying depths and tested on six datasets. The boxed area indicates models tested on unseen chaining depths.}
\label{tab:chain_exp_res}
\end{table}

\paragraph{Results.} The results are shown in Table~\ref{tab:chain_exp_res}. We can see that the models achieve high F1 scores on the respective test sets for Depth 0. The red borderline indicates F1 scores for models tested on chaining depths higher than the ones they have been trained on. We see that {Mod3} and {Mod4} do fairly well on Depth 5. However, there is a decrease for higher depths, possibly due to the need for more training examples in order to learn such depths.

Moreover, since we sample a chain of rules each time, it is likely that every model has been trained on certain chains of rules. This yields lower scores in the constant-depth test sets as the models are being tested on unseen rule chains.

Note that Mod0 shows a counter-intuitive increase in the F1 score for higher unseen depths. Chaining soft rules may lead to a low probability for the associated hypothesis, and thus eventually to a \textit{False} label. However, Mod0 is not trained on chaining and sees a hypothesis that requires chaining as an unsatisfied fact, thus eventually labelling it as \textit{False}, while in fact it is the chaining of the soft rules that is the cause for this label. This is never the case with hard rules, as the actual label there would be \textit{True}. 

\begin{table*}
\centering
\small
\begin{tabular}{llccc}
\toprule
& \bf Rule   &  \bf  \textit{FT-PLM} & \bf \textit{\systemsmall} & \bf \textit{FT-\systemsmall}  \\ 
\midrule
\multirow{5}{*}{{Known predicates}} 
& child(a,b) $\rightarrow$ parent(b,a)  & .719  & .869  & .989      \\

& relative(a,b) $\rightarrow$ negspouse(b,a) & .885 & .885 & .963     \\

& child(a,b) $\wedge$ child(b,c) $\rightarrow$ negchild(a,c) & .835  & .888 & .918    \\ 

& parent(a,b) $\wedge$ parent(a,c) $\rightarrow$ spouse(b,c)  & .754 & .757 & .814    \\ 

& parent(a,b) $\rightarrow$ negchild(a,b)   & .923 & .933 & .963     \\

\midrule
\multirow{5}{*}{{Unknown predicates}}   

& knownFor(b,a) $\rightarrow$ founder(a,b)   & .817  & .795  & .971    \\

& worksFor(b,a) $\rightarrow$ negfounder(a,b)  & .951  & .915 & .952     \\

& occupation(a, b) $\rightarrow$ negalmaMater(a, b)  & .939 & .917 & .972    \\

& author(c,b) $\wedge$ series(a,c) $\rightarrow$ author(a,b)   & .965 & .937 & .989    \\

& city(a,b) $\rightarrow$ negstate(a,b)  & .923 & .912 & .971    \\ 
\bottomrule
\end{tabular}
\caption{\label{tab:new_rules} Evaluation on unseen rules (accuracy). The first group contains rules with predicates seen by \system among the 20 rules used for fine-tuning, while the second group has rules with unseen predicates.}
\end{table*}

\subsection{Testing \system on Unseen Rules} 
\label{sec:new_rules_exp}

\begin{figure}[h]
    \centering
\small
\noindent\fbox{%
    \parbox{0.97\columnwidth}{
    
    child(a,b) $\rightarrow$ negparent(a,b)
    
    child(a,b) $\rightarrow$ nespouse(a,b)
    
    child(a,b) $\rightarrow$ negchild(b,a)
    
    child(a,b) $\rightarrow$ negrelation(b,a)

    parent(a,b) $\rightarrow$ negparent(b,a)

    parent(a,b) $\rightarrow$ nespouse(a,b)
    
    spouse(a,b) $\rightarrow$ relative(b,a)
    
    successor(a,b) $\rightarrow$ predecessor(b,a)

    predecessor(a,b) $\rightarrow$ negsuccessor(a,b)
    
    successor(a,b) $\rightarrow$ negspouse(a,b)

    predecessor(a,b) $\rightarrow$ negspouse(a,b)

    child(a,c) $\wedge$ parent(c,b) $\rightarrow$ spouse(a,b)
    
    child(b,a) $\wedge$ child(c,a) $\rightarrow$ spouse(b,c)

    parent(a,b) $\wedge$ parent(b,c) $\rightarrow$ negparent(a,c)
    
    parent(a,b) $\wedge$ child(c,a) $\rightarrow$ spouse(b,c)

    spouse(a,b) $\wedge$ parent(c,a) $\rightarrow$ negspouse(b,c)
    
    spouse(a,b) $\wedge$ child(a,c) $\rightarrow$ negspouse(b,c)

    successor(a,c) $\wedge$ successor(b,c) $\rightarrow$ negspouse(a,b)

    publisher(c,b) $\wedge$ subsequentwork(c,a) $\rightarrow$ publisher(a,b)
    
    publisher(c,b) $\wedge$ previouswork(c,a) $\rightarrow$ publisher(a,b)
    \vspace{1ex}
    }
}
    \caption{The 20 random rules used for \systemsmall.}
    \label{fig:multi_rules}
\end{figure}

We have seen that a PLM can be successfully fine-tuned with rules. We now study the performance on the PLM after it has been fine-tuned on 161 (single) rules. We call this fine-tuned model \system. 

We first evaluate \system on unseen rules. We fine-tune it with only twenty randomly selected rules (shown in Figure~\ref{fig:multi_rules}) and call it \systemsmall. We then select ten new rules divided into two groups: (\emph{i})~five rules containing predicates that were used in the rules for fine-tuning \systemsmall, and (\emph{ii})~five rules that share no predicates with the fine-tuning rules. For each rule in the test sets, we run a model fine-tuned (with 4k examples) only for that rule (FT-PLM), the model fine-tuned on the twenty original rules (\systemsmall), and the same model fine-tuned again for the rule at hand (FT-\systemsmall).

\paragraph{Results.} Table~\ref{tab:new_rules} shows that \systemsmall outperforms the fine-tuned model (FT-PLM) on the first group. Even though fine-tuned on 20 rules, it learned enough about (\emph{i})~symmetric/transitive predicates and (\emph{ii})~rule confidence to predict correctly, even better than rule-specific models. 

For the second rule group, the accuracy of \systemsmall is high, but FT-PLM performs better. Applying the same fine-tuning on \systemsmall yields the best results in all scenarios.

\section{\system on External Datasets}
\label{sec:neg_lama}

As our fine-tuning propagates information in the layers of the encoder, we hypothesize that \system effectively ``learns'' logical properties of the concepts represented in the rules, such as negation and symmetry, and thus it could perform better on tasks testing such properties of PLMs. 
To study the negation of predicates, we use the \textit{Negated LAMA datasets}, which test how PLMs distinguish a Cloze question and its negation~\citep{kassner-schutze-2020-negated}. In most cases, PLMs make the same prediction both for a positive statement (``\emph{Relativity was developed by Einstein.}'') and for its negation (``\emph{Relativity was not developed by Einstein.}''). 
To test the symmetry relationship between predicates, we use the SRL test in \textit{CheckList}~\cite{ribeiro-etal-2020-beyond}, which focuses on behavioral testing of NLP models; we use its test set for the duplicate-question detection task (QQP)~\cite{wang-etal-2018-glue}.
Finally, we test deductive reasoning on the \textit{bAbI} dataset and its Task \#15~\citep{babi}. 

\subsection{Negated LAMA Experiments}

For Negated LAMA, we do not fine-tune \system for the task; instead, we replace its original classification layer by an MLM head with weights identical to those of RoBERTa (not fine-tuned). Note that this configuration is biased in favor of RoBERTa, as the parameters of the MLM head and of the RoBERTa encoder have been trained in conjunction and thus good values have been found for this combination, which is not the case for our \system. 

\paragraph{Results} Yet, even in this arguably unfair setting, \system outperforms RoBERTa on all datasets of Negated LAMA, as shown in Table~\ref{tab:negated_lama_res}. We can see that \system performs better on both evaluation measures used in~\citep{kassner-schutze-2020-negated}. It achieves a lower mean Spearman rank correlation ($\rho$) and a much smaller percentage of positive and negated answers overlap ($\%$). 

\begin{table}[t]
\centering
\small
\begin{tabular}{l@{ }@{ }c@{ }@{ }@{ }@{ }c@{ }@{ }@{ }@{ }c}
\toprule
& \bf Fine-Tuned
& \bf RoBERTa & \bf \system \\ 

\hline
\multirow{3}{*}{bAbI} &1 epoch & .401	& .477     \\ 
& 2 epochs & .676 &	\textbf{.863} \\
& 3 epochs & .827& .825\\
\hline
{Neg. LAMA} & -  & .684 & \textbf{.852}      \\ 
\hline
\multirow{2}{*}{CheckList QQP} & 1 epoch & .000 & \textbf{.422} \\
& 3 epochs & .000 & .000 \\
\bottomrule
\end{tabular}
\caption{Evaluation on external datasets (accuracy).}
\label{tab:external}
\end{table}

\subsection{CheckList QQP Experiments} 

The CheckList tests~\citep{ribeiro-etal-2020-beyond} have shown that PLMs fail in many basic cases. We hypothesize that \system can perform better on tasks and examples that deal with symmetric and asymmetric predicates, if such predicates have been shown to it during pre-fine-tuning.
We experiment with the QQP dataset, which asks to detect whether two questions are duplicates. 
We identify a few rules that can teach a model about symmetric predicates, and we pre-fine-tune \system on them; then, we fine-tune it on the QQP dataset. 

\paragraph{Results} Table~\ref{tab:external} shows the results on the challenging CheckList QQP test set: we can see that \system achieves accuracy of 0.422 after one epoch, while RoBERTa is at 0.0. However, after three epochs \system is also at 0.0,\footnote{On the much easier QQP test set, \system achieved 0.89 accuracy after one epoch, and 0.91 after three epochs.} i.e.,~it started to unlearn what it had learned at pre-fine-tuning~\cite{forget_nn,forget_nn_2,biesialska-etal-2020-continual}. Learning a new task often leads to such catastrophic forgetting~\cite{ke-etal-2021-adapting}. While there are ways to alleviate this~\cite{ke-etal-2021-adapting}, this is beyond the scope of this paper.

\subsection{bAbI Task \#15 Experiments} 

Finally, we experiment with task \#15 of the bAbI dataset, where the goal is to assess whether a model can perform deductive reasoning. However, as mentioned in the original bAbI paper~\cite{babi}, it is not only desirable to perform well on the task, but also to use the fewest examples. 

Thus, we use the smallest dataset consisting of about 2,000 data points. We hypothesize that under the same conditions and hyper-parameters, \system should be able to generalize faster and to learn in fewer epochs. As PLMs produce varying scores when fine-tuned on small datasets, we repeat the experiment ten times and we report the average scores. We then compare to RoBERTa. Both models contain two classification layers to predict start and end spans of the input context.

\begin{table}[t]
\centering
\small
\label{tab:neg_lama}
\begin{tabular}{@{}l@{}c@{}r@{ }|@{ }r@{ }@{ }r@{ }|@{ }r@{ }r@{}}
\toprule
&&\bf Facts 
& \multicolumn{2}{c}{\bf RoBERTa} & \multicolumn{2}{c}{\bf \system}    \\ 

&&&$\rho$&\%&$\rho$&\%  \\ 
 
\hline
\multirow{3}{*}{GR} &birthplace&2,404 &90.99&18.51&71.72&4.20      \\ 
& birthdate&1,565 &82.87&1.40&63.55&0.13 \\
& deathplace&649 &86.44&0.31&71.13&0.00 \\

\hline
\multirow{3}{*}{T-REx} &1-1&973 &78.95&61.38&51.21&32.96      \\ 
& N-1&20,006 &87.56&43.80&67.63&11.48 \\
& N-M&13,096 &89.39&50.78&72.59&28.90 \\
\hline

ConceptNet&---&2,996& 42.61&9.00&37.43&4.83 \\

SQ&---&286& 89.71&44.76&75.05&26.32 \\
\bottomrule
\end{tabular}
\caption{Negated LAMA: Mean Spearman rank correlation $(\rho)$ and mean percentage of overlap in the first ranked predictions (\%) for original vs. negated queries.}
\label{tab:negated_lama_res}
\end{table}

\paragraph{Results} We can see in Table~\ref{tab:external} that \system achieves accuracy of 0.863 in two epochs, while RoBERTa achieves 0.676. On the third epoch, RoBERTa catches up with accuracy of 0.827, while \system starts to overfit (goes down to 0.825), indicating that fewer epochs should be used, probably due to catastrophic forgetting.

\section{Conclusion and Future Work}
\label{sec:conclusion}

We studied whether PLMs could reason with soft rules over natural language. We experimented with one flavor of probabilistic answer set programming (\lpmln), but other semantics can be also used with the proposed methodology. We further explored the inference capabilities of Transformer-based PLMs, focusing on positive and negative textual entailment. 

We leave non-entailment for future work. We also leave open the development of explainable models. Some approaches use occlusion that removes parts of the input and checks the impact on the output~\citep{ruletakers} or build proof iteratively using 1-hop inference~\citep{tafjord2020proofwriter}.

\section*{Acknowledgments}

This work is partially supported by a Google Faculty Research Award and the ANR JCJC Grant \textit{InfClean}.

\section*{Ethics and Broader Impact}

\paragraph{Data Collection}
While we generated the facts in our examples, the logical rules have been mined from the data in the DBpedia knowledge graph, which in turn has been generated from Wikipedia.

\paragraph{Biases}
We are aware of (\emph{i})~the biases and abusive language patterns~\cite{sheng-etal-2019-woman,Zhang2020,stochatic_parrots,liang2021towards} that PLMs impose, and (\emph{ii})~the imperfectness and the biases of our rules as data from Wikipedia has been used to mine the rules and compute their confidences~~\cite{Janowicz2018DebiasingKG,BiasKG}. However, our goal is to study PLM's capability of deductive soft reasoning. For (\emph{i}), there has been some work on debiasing PLMs~\cite{liang-etal-2020-towards}, while for (\emph{ii}), we used mined rules to have more variety, but could resort to user-specified rules validated by consensus to relieve the bias.

\paragraph{Environmental Impact}

The use of large-scale Transformers requires a lot of computations and GPUs/TPUs for training, which contributes to global warming~\cite{strubell-etal-2019-energy,GreenAI}. This is a smaller issue in our case, as we do not train such models from scratch; rather, we fine-tune them on relatively small datasets. Moreover, running on a CPU for inference, once the model is fine-tuned, is less problematic as CPUs have a much lower environmental impact.

\bibliographystyle{acl_natbib}
\bibliography{rulebert}

\begin{thebibliography}{49}
\expandafter\ifx\csname natexlab\endcsname\relax\def\natexlab#1{#1}\fi

\bibitem[{Ahmadi et~al.(2020)Ahmadi, Truong, Dao, Ortona, and
  Papotti}]{AhmadiTDOP20}
Naser Ahmadi, Thi-Thuy-Duyen Truong, Le-Hong-Mai Dao, Stefano Ortona, and Paolo
  Papotti. 2020.
\newblock \href {https://doi.org/10.1145/3409384} {{RuleHub}: A public corpus
  of rules for knowledge graphs}.
\newblock \emph{J. Data and Information Quality}, 12(4).

\bibitem[{Baral(2010)}]{chittaBook}
Chitta Baral. 2010.
\newblock \emph{Knowledge Representation, Reasoning and Declarative Problem
  Solving}, 1st edition.
\newblock Cambridge University Press, USA.

\bibitem[{Bender et~al.(2021)Bender, Gebru, McMillan-Major, and
  Shmitchell}]{stochatic_parrots}
Emily~M. Bender, Timnit Gebru, Angelina McMillan-Major, and Shmargaret
  Shmitchell. 2021.
\newblock \href {https://doi.org/10.1145/3442188.3445922} {On the dangers of
  stochastic parrots: Can language models be too big?}
\newblock In \emph{Proceedings of the 2021 ACM Conference on Fairness,
  Accountability, and Transparency}, FAccT '21, page 610–623, Virtual Event,
  Canada. Association for Computing Machinery.

\bibitem[{Biesialska et~al.(2020)Biesialska, Biesialska, and
  Costa-juss{\`a}}]{biesialska-etal-2020-continual}
Magdalena Biesialska, Katarzyna Biesialska, and Marta~R. Costa-juss{\`a}. 2020.
\newblock \href {https://doi.org/10.18653/v1/2020.coling-main.574} {Continual
  lifelong learning in natural language processing: A survey}.
\newblock In \emph{Proceedings of the 28th International Conference on
  Computational Linguistics}, COLING~'20, pages 6523--6541, Barcelona, Spain
  (Online). International Committee on Computational Linguistics.

\bibitem[{Bisong(2019)}]{Bisong2019}
Ekaba Bisong. 2019.
\newblock \href {https://doi.org/10.1007/978-1-4842-4470-8_7} {Google
  colaboratory}.
\newblock In \emph{Building Machine Learning and Deep Learning Models on Google
  Cloud Platform: A Comprehensive Guide for Beginners}, pages 59--64. Apress.

\bibitem[{Clark et~al.(2019{\natexlab{a}})Clark, Lee, Chang, Kwiatkowski,
  Collins, and Toutanova}]{clark2019boolq}
Christopher Clark, Kenton Lee, Ming-Wei Chang, Tom Kwiatkowski, Michael
  Collins, and Kristina Toutanova. 2019{\natexlab{a}}.
\newblock \href {https://doi.org/10.18653/v1/N19-1300} {{B}ool{Q}: Exploring
  the surprising difficulty of natural yes/no questions}.
\newblock In \emph{Proceedings of the 2019 Conference of the North {A}merican
  Chapter of the Association for Computational Linguistics: Human Language
  Technologies, Volume 1 (Long and Short Papers)}, NAACL-HLT~'19, pages
  2924--2936, Minneapolis, Minnesota, USA. Association for Computational
  Linguistics.

\bibitem[{Clark et~al.(2019{\natexlab{b}})Clark, Khandelwal, Levy, and
  Manning}]{clark2019what}
Kevin Clark, Urvashi Khandelwal, Omer Levy, and Christopher~D. Manning.
  2019{\natexlab{b}}.
\newblock \href {https://doi.org/10.18653/v1/W19-4828} {What does {BERT} look
  at? {A}n analysis of {BERT}{'}s attention}.
\newblock In \emph{Proceedings of the 2019 ACL Workshop BlackboxNLP: Analyzing
  and Interpreting Neural Networks for NLP}, BlackboxNLP~'19, pages 276--286,
  Florence, Italy. Association for Computational Linguistics.

\bibitem[{Clark et~al.(2020)Clark, Tafjord, and Richardson}]{ruletakers}
Peter Clark, Oyvind Tafjord, and Kyle Richardson. 2020.
\newblock \href {https://doi.org/10.24963/ijcai.2020/537} {Transformers as soft
  reasoners over language}.
\newblock In \emph{Proceedings of the Twenty-Ninth International Joint
  Conference on Artificial Intelligence,}, IJCAI~'20, pages 3882--3890, Online.
  International Joint Conferences on Artificial Intelligence Organization.

\bibitem[{Dagan et~al.(2013)Dagan, Roth, Sammons, and Zanzotto}]{2013Dagan}
Ido Dagan, Dan Roth, Mark Sammons, and Fabio~Massimo Zanzotto. 2013.
\newblock \emph{{R}ecognizing {T}extual {E}ntailment: {M}odels and
  {A}pplications}.
\newblock Synthesis Lectures on Human Language Technologies. Morgan and
  Claypool publishers.

\bibitem[{Demartini(2019)}]{BiasKG}
Gianluca Demartini. 2019.
\newblock \href {https://doi.org/10.1145/3308560.3317307} {Implicit bias in
  crowdsourced knowledge graphs}.
\newblock In \emph{Proceedings of the 2019 World Wide Web Conference: Companion
  Volume}, WWW~'19, page 624–630, San Francisco, California, USA. Association
  for Computing Machinery.

\bibitem[{Devlin et~al.(2019)Devlin, Chang, Lee, and
  Toutanova}]{devlin-etal-2019-bert}
Jacob Devlin, Ming-Wei Chang, Kenton Lee, and Kristina Toutanova. 2019.
\newblock \href {https://doi.org/10.18653/v1/N19-1423} {{BERT}: Pre-training of
  deep bidirectional transformers for language understanding}.
\newblock In \emph{Proceedings of the 2019 Conference of the North {A}merican
  Chapter of the Association for Computational Linguistics: Human Language
  Technologies, Volume 1 (Long and Short Papers)}, NAACL-HLT~'19, pages
  4171--4186, Minneapolis, Minnesota, USA. Association for Computational
  Linguistics.

\bibitem[{Dodge et~al.(2020)Dodge, Ilharco, Schwartz, Farhadi, Hajishirzi, and
  Smith}]{dodge2020finetuning}
Jesse Dodge, Gabriel Ilharco, Roy Schwartz, Ali Farhadi, Hannaneh Hajishirzi,
  and Noah~A. Smith. 2020.
\newblock \href {https://arxiv.org/abs/2002.06305} {Fine-tuning pretrained
  language models: Weight initializations, data orders, and early stopping}.
\newblock \emph{arXiv:2002.06305}.

\bibitem[{Elazar et~al.(2021)Elazar, Kassner, Ravfogel, Ravichander, Hovy,
  Sch{\"{u}}tze, and Goldberg}]{Elazar2021MeasuringAI}
Yanai Elazar, Nora Kassner, Shauli Ravfogel, Abhilasha Ravichander, Eduard~H.
  Hovy, Hinrich Sch{\"{u}}tze, and Yoav Goldberg. 2021.
\newblock \href {https://arxiv.org/abs/2102.01017} {Measuring and improving
  consistency in pretrained language models}.
\newblock \emph{arXiv:2102.01017}.

\bibitem[{Gal\'{a}rraga et~al.(2015)Gal\'{a}rraga, Teflioudi, Hose, and
  Suchanek}]{galarraga2015fast}
Luis Gal\'{a}rraga, Christina Teflioudi, Katja Hose, and Fabian~M. Suchanek.
  2015.
\newblock \href {https://doi.org/10.1007/s00778-015-0394-1} {Fast rule mining
  in ontological knowledge bases with {AMIE++}}.
\newblock \emph{The VLDB Journal}, 24(6):707–730.

\bibitem[{Gebser et~al.(2014)Gebser, Kaminski, Kaufmann, and
  Schaub}]{GebserKKS14}
Martin Gebser, Roland Kaminski, Benjamin Kaufmann, and Torsten Schaub. 2014.
\newblock \href {http://arxiv.org/abs/1405.3694} {Clingo = {ASP} + control:
  Preliminary report}.
\newblock \emph{arXiv:1405.3694}.

\bibitem[{Hamilton et~al.(2018)Hamilton, Bajaj, Zitnik, Jurafsky, and
  Leskovec}]{hamilton-2018-nips}
William~L. Hamilton, Payal Bajaj, Marinka Zitnik, Dan Jurafsky, and Jure
  Leskovec. 2018.
\newblock \href {https://dl.acm.org/doi/10.5555/3326943.3327131} {Embedding
  logical queries on knowledge graphs}.
\newblock In \emph{Proceedings of the 32nd International Conference on Neural
  Information Processing Systems}, NIPS'18, page 2030–2041, Montr\'{e}al,
  Canada. Curran Associates Inc.

\bibitem[{Janowicz et~al.(2018)Janowicz, Yan, Regalia, Zhu, and
  Mai}]{Janowicz2018DebiasingKG}
Krzysztof Janowicz, Bo~Yan, Blake Regalia, Rui Zhu, and Gengchen Mai. 2018.
\newblock \href
  {http://ceur-ws.org/Vol-2180/ISWC_2018_Outrageous_Ideas_paper_17.pdf}
  {Debiasing knowledge graphs: Why female presidents are not like female
  popes}.
\newblock In \emph{Proceedings of the International Semantic Web Conference},
  ISWC~'18, Monterey, California, USA.

\bibitem[{Kassner et~al.(2020)Kassner, Krojer, and
  Sch{\"u}tze}]{kassner-etal-2020-pretrained}
Nora Kassner, Benno Krojer, and Hinrich Sch{\"u}tze. 2020.
\newblock \href {https://doi.org/10.18653/v1/2020.conll-1.45} {Are pretrained
  language models symbolic reasoners over knowledge?}
\newblock In \emph{Proceedings of the 24th Conference on Computational Natural
  Language Learning}, CoNLL~'20, pages 552--564, Online. Association for
  Computational Linguistics.

\bibitem[{Kassner and Sch{\"u}tze(2020)}]{kassner-schutze-2020-negated}
Nora Kassner and Hinrich Sch{\"u}tze. 2020.
\newblock \href {https://doi.org/10.18653/v1/2020.acl-main.698} {Negated and
  misprimed probes for pretrained language models: Birds can talk, but cannot
  fly}.
\newblock In \emph{Proceedings of the 58th Annual Meeting of the Association
  for Computational Linguistics}, ACL~'20, pages 7811--7818, Online.
  Association for Computational Linguistics.

\bibitem[{Ke et~al.(2021)Ke, Xu, and Liu}]{ke-etal-2021-adapting}
Zixuan Ke, Hu~Xu, and Bing Liu. 2021.
\newblock \href {https://doi.org/10.18653/v1/2021.naacl-main.378} {Adapting
  {BERT} for continual learning of a sequence of aspect sentiment
  classification tasks}.
\newblock In \emph{Proceedings of the 2021 Conference of the North American
  Chapter of the Association for Computational Linguistics: Human Language
  Technologies}, NAACL-HLT~'21, pages 4746--4755, Online. Association for
  Computational Linguistics.

\bibitem[{Kemker et~al.(2018)Kemker, McClure, Abitino, Hayes, and
  Kanan}]{forget_nn_2}
Ronald Kemker, Marc McClure, Angelina Abitino, Tyler~L. Hayes, and Christopher
  Kanan. 2018.
\newblock \href
  {https://www.aaai.org/ocs/index.php/AAAI/AAAI18/paper/view/16410} {Measuring
  catastrophic forgetting in neural networks}.
\newblock In \emph{Proceedings of the Thirty-Second {AAAI} Conference on
  Artificial Intelligence}, AAAI~'18, pages 3390--3398, New Orleans, Louisiana,
  USA. {AAAI} Press.

\bibitem[{Kirkpatrick et~al.(2017)Kirkpatrick, Pascanu, Rabinowitz, Veness,
  Desjardins, Rusu, Milan, Quan, Ramalho, Grabska-Barwinska, Hassabis, Clopath,
  Kumaran, and Hadsell}]{forget_nn}
James Kirkpatrick, Razvan Pascanu, Neil Rabinowitz, Joel Veness, Guillaume
  Desjardins, Andrei~A. Rusu, Kieran Milan, John Quan, Tiago Ramalho, Agnieszka
  Grabska-Barwinska, Demis Hassabis, Claudia Clopath, Dharshan Kumaran, and
  Raia Hadsell. 2017.
\newblock \href {https://doi.org/10.1073/pnas.1611835114} {{O}vercoming
  catastrophic forgetting in neural networks}.
\newblock \emph{Proceedings of the National Academy of Sciences},
  114(13):3521--3526.

\bibitem[{Lample and Charton(2020)}]{Lample2020Deep}
Guillaume Lample and Fran{\c{c}}ois Charton. 2020.
\newblock \href {https://openreview.net/forum?id=S1eZYeHFDS} {Deep learning for
  symbolic mathematics}.
\newblock In \emph{Proceedings of the 8th International Conference on Learning
  Representations}, ICLR~'20, Addis Ababa, Ethiopia. OpenReview.net.

\bibitem[{Lee et~al.(2017)Lee, Talsania, and Wang}]{Lee2017ComputingLU}
Joohyung Lee, Samidh Talsania, and Y.~Wang. 2017.
\newblock \href {https://doi.org/10.1017/S1471068417000400} {Computing {LPMLN}
  using {ASP} and {MLN} solvers}.
\newblock \emph{Theory and Practice of Logic Programming}, 17(5--6):942--960.

\bibitem[{Lee and Wang(2016)}]{lee2016weighted}
Joohyung Lee and Yi~Wang. 2016.
\newblock \href {https://dl.acm.org/doi/10.5555/3032027.3032045} {Weighted
  rules under the stable model semantics}.
\newblock In \emph{Proceedings of the Fifteenth International Conference on
  Principles of Knowledge Representation and Reasoning}, KR~'16, page
  145–154, Cape Town, South Africa. AAAI Press.

\bibitem[{Liang et~al.(2020)Liang, Li, Zheng, Lim, Salakhutdinov, and
  Morency}]{liang-etal-2020-towards}
Paul~Pu Liang, Irene~Mengze Li, Emily Zheng, Yao~Chong Lim, Ruslan
  Salakhutdinov, and Louis-Philippe Morency. 2020.
\newblock \href {https://doi.org/10.18653/v1/2020.acl-main.488} {Towards
  debiasing sentence representations}.
\newblock In \emph{Proceedings of the 58th Annual Meeting of the Association
  for Computational Linguistics}, ACL~'20, pages 5502--5515, Online.
  Association for Computational Linguistics.

\bibitem[{Liang et~al.(2021)Liang, Wu, Morency, and
  Salakhutdinov}]{liang2021towards}
Paul~Pu Liang, Chiyu Wu, Louis-Philippe Morency, and Ruslan Salakhutdinov.
  2021.
\newblock \href {https://arxiv.org/abs/2106.13219} {Towards understanding and
  mitigating social biases in language models}.
\newblock In \emph{Proceedings of the International Conference on Machine
  Learning}, ICML~'21, pages 6565--6576, Online. PMLR.

\bibitem[{Liang(2016)}]{liang-2016}
Percy Liang. 2016.
\newblock \href {https://doi.org/10.1145/2866568} {Learning executable semantic
  parsers for natural language understanding}.
\newblock \emph{Commun. ACM}, 59(9):68–76.

\bibitem[{Liu et~al.(2020)Liu, Ott, Goyal, Du, Joshi, Chen, Levy, Lewis,
  Zettlemoyer, and Veselin}]{roberta2020}
Yinhan Liu, Myle Ott, Naman Goyal, Jingfei Du, Mandar Joshi, Danqi Chen, Omer
  Levy, Mike Lewis, Luke Zettlemoyer, and Veselin. 2020.
\newblock \href {https://openreview.net/forum?id=SyxS0T4tvS} {{RoBERTa}: A
  robustly optimized {BERT} pretraining approach}.
\newblock In \emph{Proceedings of the 8th International Conference on Learning
  Representations}, ICLR~'20, Addis Ababa, Ethiopia. OpenReview.net.

\bibitem[{MacCartney and Manning(2009)}]{maccartney-manning-2009-extended}
Bill MacCartney and Christopher~D. Manning. 2009.
\newblock \href {https://aclanthology.org/W09-3714} {An extended model of
  natural logic}.
\newblock In \emph{Proceedings of the Eight International Conference on
  Computational Semantics}, IWCS-WS~'09, pages 140--156, Tilburg, The
  Netherlands. Association for Computational Linguistics.

\bibitem[{Minervini et~al.(2020)Minervini, Bošnjak, Rocktäschel, Riedel, and
  Grefenstette}]{Minervini_2020}
Pasquale Minervini, Matko Bošnjak, Tim Rocktäschel, Sebastian Riedel, and
  Edward Grefenstette. 2020.
\newblock \href {https://doi.org/10.1609/aaai.v34i04.5962} {Differentiable
  reasoning on large knowledge bases and natural language}.
\newblock \emph{Proceedings of the AAAI Conference on Artificial Intelligence},
  34(04):5182--5190.

\bibitem[{Mosbach et~al.(2021)Mosbach, Andriushchenko, and
  Klakow}]{mosbach2021on}
Marius Mosbach, Maksym Andriushchenko, and Dietrich Klakow. 2021.
\newblock \href {https://openreview.net/forum?id=nzpLWnVAyah} {On the stability
  of fine-tuning bert: Misconceptions, explanations, and strong baselines}.
\newblock In \emph{Proceedings of the 9th International Conference on Learning
  Representations}, ICLR~'21, Virtual Event, Austria. OpenReview.net.

\bibitem[{Ortona et~al.(2018)Ortona, Meduri, and Papotti}]{OrtonaMP18}
Stefano Ortona, Venkata~Vamsikrishna Meduri, and Paolo Papotti. 2018.
\newblock \href {https://doi.org/10.1109/ICDE.2018.00108} {Robust discovery of
  positive and negative rules in knowledge bases}.
\newblock In \emph{Proceedings of the 2018 IEEE 34th International Conference
  on Data Engineering}, ICDE~'18, pages 1168--1179, Paris, France. IEEE.

\bibitem[{Petroni et~al.(2019)Petroni, Rockt{\"a}schel, Riedel, Lewis, Bakhtin,
  Wu, and Miller}]{petroni-etal-2019-language}
Fabio Petroni, Tim Rockt{\"a}schel, Sebastian Riedel, Patrick Lewis, Anton
  Bakhtin, Yuxiang Wu, and Alexander Miller. 2019.
\newblock \href {https://doi.org/10.18653/v1/D19-1250} {Language models as
  knowledge bases?}
\newblock In \emph{Proceedings of the 2019 Conference on Empirical Methods in
  Natural Language Processing and the 9th International Joint Conference on
  Natural Language Processing}, EMNLP-IJCNLP~'19, pages 2463--2473, Hong Kong,
  China. Association for Computational Linguistics.

\bibitem[{Ribeiro et~al.(2020)Ribeiro, Wu, Guestrin, and
  Singh}]{ribeiro-etal-2020-beyond}
Marco~Tulio Ribeiro, Tongshuang Wu, Carlos Guestrin, and Sameer Singh. 2020.
\newblock \href {https://doi.org/10.18653/v1/2020.acl-main.442} {Beyond
  accuracy: Behavioral testing of {NLP} models with {C}heck{L}ist}.
\newblock In \emph{Proceedings of the 58th Annual Meeting of the Association
  for Computational Linguistics}, ACL~'20, pages 4902--4912, Online.
  Association for Computational Linguistics.

\bibitem[{Rogers et~al.(2020)Rogers, Kovaleva, and
  Rumshisky}]{rogers-etal-2020-primer}
Anna Rogers, Olga Kovaleva, and Anna Rumshisky. 2020.
\newblock \href {https://doi.org/10.1162/tacl_a_00349} {A primer in
  {BERT}ology: What we know about how {BERT} works}.
\newblock \emph{Transactions of the Association for Computational Linguistics},
  8:842--866.

\bibitem[{Schwartz et~al.(2020)Schwartz, Dodge, Smith, and Etzioni}]{GreenAI}
Roy Schwartz, Jesse Dodge, Noah~A. Smith, and Oren Etzioni. 2020.
\newblock \href {https://doi.org/10.1145/3381831} {Green {AI}}.
\newblock \emph{Commun. ACM}, 63(12):54–63.

\bibitem[{Sheng et~al.(2019)Sheng, Chang, Natarajan, and
  Peng}]{sheng-etal-2019-woman}
Emily Sheng, Kai-Wei Chang, Premkumar Natarajan, and Nanyun Peng. 2019.
\newblock \href {https://doi.org/10.18653/v1/D19-1339} {The woman worked as a
  babysitter: On biases in language generation}.
\newblock In \emph{Proceedings of the 2019 Conference on Empirical Methods in
  Natural Language Processing and the 9th International Joint Conference on
  Natural Language Processing}, EMNLP-IJCNLP~'19, pages 3407--3412, Hong Kong,
  China. Association for Computational Linguistics.

\bibitem[{Strubell et~al.(2019)Strubell, Ganesh, and
  McCallum}]{strubell-etal-2019-energy}
Emma Strubell, Ananya Ganesh, and Andrew McCallum. 2019.
\newblock \href {https://doi.org/10.18653/v1/P19-1355} {Energy and policy
  considerations for deep learning in {NLP}}.
\newblock In \emph{Proceedings of the 57th Annual Meeting of the Association
  for Computational Linguistics}, ACL~'19, pages 3645--3650, Florence, Italy.
  Association for Computational Linguistics.

\bibitem[{Tafjord et~al.(2021)Tafjord, Dalvi, and
  Clark}]{tafjord2020proofwriter}
Oyvind Tafjord, Bhavana Dalvi, and Peter Clark. 2021.
\newblock \href {https://doi.org/10.18653/v1/2021.findings-acl.317}
  {{P}roof{W}riter: Generating implications, proofs, and abductive statements
  over natural language}.
\newblock In \emph{Findings of the Association for Computational Linguistics},
  ACL-IJCNLP~'21, pages 3621--3634, Online. Association for Computational
  Linguistics.

\bibitem[{Talmor et~al.(2020{\natexlab{a}})Talmor, Elazar, Goldberg, and
  Berant}]{olympics}
Alon Talmor, Yanai Elazar, Yoav Goldberg, and Jonathan Berant.
  2020{\natexlab{a}}.
\newblock \href {https://doi.org/10.1162/tacl\_a\_00342} {o{LM}pics-on what
  language model pre-training captures}.
\newblock \emph{Transactions of the Association for Computational Linguistics},
  8:743--758.

\bibitem[{Talmor et~al.(2020{\natexlab{b}})Talmor, Tafjord, Clark, Goldberg,
  and Berant}]{leap_of_thought}
Alon Talmor, Oyvind Tafjord, Peter Clark, Yoav Goldberg, and Jonathan Berant.
  2020{\natexlab{b}}.
\newblock \href
  {https://proceedings.neurips.cc/paper/2020/file/e992111e4ab9985366e806733383bd8c-Paper.pdf}
  {Leap-of-thought: Teaching pre-trained models to systematically reason over
  implicit knowledge}.
\newblock In \emph{Advances in Neural Information Processing Systems 33: Annual
  Conference on Neural Information Processing Systems 2020}, volume~33 of
  \emph{NeurIPS~'20}, pages 20227--20237, Online.

\bibitem[{Wang et~al.(2019{\natexlab{a}})Wang, Singh, Michael, Hill, Levy, and
  Bowman}]{wang-etal-2018-glue}
Alex Wang, Amanpreet Singh, Julian Michael, Felix Hill, Omer Levy, and
  Samuel~R. Bowman. 2019{\natexlab{a}}.
\newblock \href {https://openreview.net/forum?id=rJ4km2R5t7} {{GLUE:} {A}
  multi-task benchmark and analysis platform for natural language
  understanding}.
\newblock In \emph{Proceedings of the 7th International Conference on Learning
  Representations}, ICLR~'19, New Orleans, Louisiana, USA. OpenReview.net.

\bibitem[{Wang et~al.(2019{\natexlab{b}})Wang, Li, Xiao, Zhu, Li, Wong, and
  Chao}]{wang-etal-2019-learning}
Qiang Wang, Bei Li, Tong Xiao, Jingbo Zhu, Changliang Li, Derek~F. Wong, and
  Lidia~S. Chao. 2019{\natexlab{b}}.
\newblock \href {https://doi.org/10.18653/v1/P19-1176} {Learning deep
  transformer models for machine translation}.
\newblock In \emph{Proceedings of the 57th Annual Meeting of the Association
  for Computational Linguistics}, ACL~'19, pages 1810--1822, Florence, Italy.
  Association for Computational Linguistics.

\bibitem[{Wang and Chen(2020)}]{wang-chen-2020-position}
Yu-An Wang and Yun-Nung Chen. 2020.
\newblock \href {https://doi.org/10.18653/v1/2020.emnlp-main.555} {What do
  position embeddings learn? {A}n empirical study of pre-trained language model
  positional encoding}.
\newblock In \emph{Proceedings of the 2020 Conference on Empirical Methods in
  Natural Language Processing}, EMNLP~'20, pages 6840--6849, Online.
  Association for Computational Linguistics.

\bibitem[{Weston et~al.(2016)Weston, Bordes, Chopra, and Mikolov}]{babi}
Jason Weston, Antoine Bordes, Sumit Chopra, and Tom{\'{a}}s Mikolov. 2016.
\newblock \href {http://arxiv.org/abs/1502.05698} {Towards {AI}-complete
  question answering: {A} set of prerequisite toy tasks}.
\newblock In \emph{Proceedings of the 4th International Conference on Learning
  Representations}, ICLR~'16, San Juan, Puerto Rico.

\bibitem[{Yang et~al.(2017)Yang, Yang, and Cohen}]{yang-2017-nips}
Fan Yang, Zhilin Yang, and William~W. Cohen. 2017.
\newblock \href
  {https://proceedings.neurips.cc/paper/2017/hash/0e55666a4ad822e0e34299df3591d979-Abstract.html}
  {Differentiable learning of logical rules for knowledge base reasoning}.
\newblock In \emph{Advances in Neural Information Processing Systems 30: Annual
  Conference on Neural Information Processing Systems 2017, December 4-9, 2017,
  Long Beach, California, {USA}}, NeurIPS~'17, pages 2319--2328.

\bibitem[{Zhang et~al.(2020)Zhang, Lu, Abdalla, McDermott, and
  Ghassemi}]{Zhang2020}
Haoran Zhang, Amy~X. Lu, Mohamed Abdalla, Matthew McDermott, and Marzyeh
  Ghassemi. 2020.
\newblock \href {https://doi.org/10.1145/3368555.3384448} {Hurtful words:
  Quantifying biases in clinical contextual word embeddings}.
\newblock In \emph{Proceedings of the ACM Conference on Health, Inference, and
  Learning}, CHIL~'20, page 110–120, Toronto, Ontario, Canada. Association
  for Computing Machinery.

\bibitem[{Zhou et~al.(2020)Zhou, Zhang, Cui, and Huang}]{cat}
Xuhui Zhou, Yue Zhang, Leyang Cui, and Dandan Huang. 2020.
\newblock \href {https://doi.org/10.1609/aaai.v34i05.6523} {Evaluating
  commonsense in pre-trained language models}.
\newblock In \emph{Proceedings of the AAAI Conference on Artificial
  Intelligence}, volume~34 of \emph{AAAI~'20}, pages 9733--9740, Online. {AAAI}
  Press.

\end{thebibliography}

\clearpage
\newpage
\appendix
\section{More on Reasoning with Soft Rules}

Let $\sigma$ be a signature as in first-order logic. An \lpmln \ program $\Pi$ is a finite set of weighted rules of the form:
\beq
w: A \leftarrow B 
\eeq{lpmln-rule}
where $A$ is a disjunction of atoms of $\sigma$, $B$ is a conjunction of literals (atoms and negated atoms) of $\sigma$, and $w$ is a real number or the symbol $\alpha$.

When $A$ is $\bot$ (the empty disjunction), the rule asserts that $B$ should be false in the stable model. An \lpmln \ rule \eqref{lpmln-rule} is called {\em soft} if $w$ is a real number or {\em hard} if $w$ is $\alpha$. An \lpmln \ program is {\em ground} if its rules contain no variables. An \lpmln \ program $\Pi$ that contains variables is identified with a ground \lpmln \ program $gr_{\sigma}[\Pi]$, which is obtained from $\Pi$ by replacing every variable with every ground term of $\sigma$. The weight of a ground rule in $gr_{\sigma}[\Pi]$ is the same as the weight of the corresponding rule in $\Pi$. By $\overline{\Pi}$ we denote the unweighted logic program obtained from $\Pi$, i.e.,~$\overline{\Pi} =\{R \hspace{0.1cm}|\hspace{0.1cm} w:R\in\Pi\}$. 

For a ground \lpmln \ program $\Pi$, ${\Pi}_I$ denotes the set of rules $w: R$ in $\Pi$ such that $I$ satisfies $R$ (denoted $I\models R$) and {\rm SM}$[\Pi]$ denotes the set $\{I \mid \text{$I$ is a (deterministic) stable model of $\overline{\Pi_I}$}\}$. The (unnormalized) weight of $I$ under $\Pi$ is defined as follows:
\[
\begin{aligned}
    W_{\Pi}(I) = 
    \begin{cases}
        exp(\sum\limits_{w:R\in\Pi_I} w) \  \text{ if } I \in \sm[\Pi];\\
        0 \hspace{4.5px} \hspace{2cm} $otherwise$.
        \end{cases}
\end{aligned}
\]
The probability of $I$ under $\Pi$ is the normalized weight defined as follows:
$$P\textsubscript{$\Pi$}(I) = \lim_{\alpha\to\infty} \frac{W\textsubscript{$\Pi$}(I)}{\sum_{J\in {\rm SM}[\Pi]} W\textsubscript{{$\Pi$}}(J)}.
$$

In Answer Set programming (ASP), search problems are reduced to computing {\em stable models} (a.k.a. answer sets), a set of beliefs described by the program. In the case of a Horn program, the stable models coincide with the minimal models. \lpmln \ programs are transformed to meet the needs of an ASP solver~\cite{GebserKKS14,Lee2017ComputingLU}.

\begin{figure}
\hspace{-2ex}
\begin{tikzpicture}
\begin{axis}[
    ymin=0,
    ybar,
    width=8.0cm,
    height=5cm,
    enlargelimits=0.15,
    legend style={at={(0.63,1.03)},
     anchor=north,legend columns=-1},
    ylabel={Average Support},
    xlabel = {Number of rules},
    symbolic x coords={1,2,3,4,5},%,6,7,8,9,10},
    xtick=data,
    nodes near coords align={vertical},
    ]

\addplot coordinates { 
(1,8852) (2,4309) (3,1962) (4,1098) (5,591)};

\addplot coordinates { 
(1,34436) (2,15540) (3,5126) (4,1486) (5,319)};

\addplot coordinates { 
(1,5725) (2,1185) (3,169) (4,6) (5,0)};

\legend{spouse,child,relative}
\end{axis}
\end{tikzpicture}

    \caption{Support of the overlapping rules.}
    \label{fig:supports}

\end{figure}
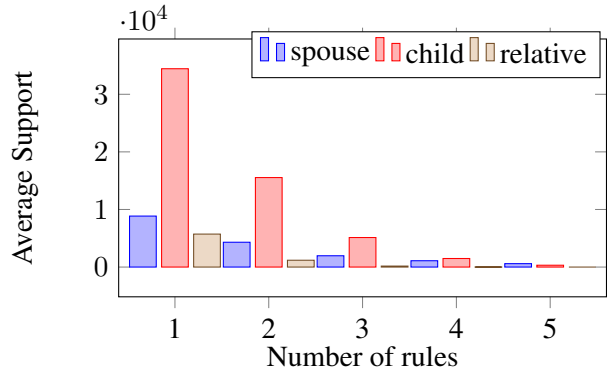

\section{Rule Support}

We designed an experiment to show the impact of increasing the number of overlapping rules on the same target predicate. The goal is to measure how often multiple rules are triggered for the same target triple.

We measure this with the support of a rule, i.e.,~the number of triples in the knowledge base that satisfy all the atoms in the rule. 

To compute the support for more than one rule, we combine the premises of the rules. In this experiment, we picked three predicates (\textit{spouse}, \textit{child} and \textit{relative}), and for each one we selected ten rules randomly. Next, we used DBpedia online endpoint\footnote{\url{http://dbpedia.org/sparql}} to compute the support for each combination of $n$ ($n$=1,2,...,5) rules for each predicate. The results in Figure~\ref{fig:supports} show that by increasing the number of rules, the support decreases for all predicates. For combinations with more than three rules, the support is very small.  

\section{More Experimental Details}
\label{sec:exps_details}

For fine-tuning our models, we use Google Colaboratory~\cite{Bisong2019}, which assigns random GPU clusters of various types. The number of parameters of our models is about 355M. We select the values of our hyper-parameters (shown in Table~\ref{tab:fine_tune_hyperparam}) on the development sets, by maximizing accuracy.

The execution times vary largely depending on the GPU at hand and on the scenario, with fine-tuning on a Tesla V100 taking from one hour for a single rule to a few hours for all the chaining experiments. The training/validation/testing splits are shown in Table~\ref{tab:dataset_distributions}. Table~\ref{tab:chain_dataset_sizes} shows the sizes of the used test datasets.

\begin{table}[tbh]
    \centering
    \small
    \begin{tabular}{lr}
        \toprule
        \textbf{Hyper-Parameter} & \textbf{Value}  \\
        \midrule
         Learning Rate & 1e-6\\
         Weight Decay & 0.1\\
         Number of Epochs & 3\\
         Batch Size & 16\\
         Learning Rate Decay & Linear\\
         Warmup Ratio & 0.06\\
         \bottomrule
    \end{tabular}
    \caption{Hyper-parameters for fine-tuning our model.}
    \label{tab:fine_tune_hyperparam}
\end{table}

\begin{table}[tbh]
    \centering
    \small
    \begin{tabular}{lc}
        \toprule
         \bf Dataset & \bf Size \\
         \midrule
         Mod0 Test(own)  &  2,667\\
         Mod1 Test(own)  &  4,000 \\
         Mod2 Test(own)  & 5,334 \\
         Mod3 Test(own)  & 6,667  \\
         Mod4 Test(own)  & 8,000  \\
         Mod5 Test(own)  & 9,334  \\
        Test(D$\leq$5)& 9,334
        \\
         Depth=0    &16,057
        \\
        Depth=1    & 6,608
        \\ 
        Depth=2    & 5,389
        \\
        Depth=3    & 3,993
        \\
        Depth=4    & 2,619
        \\
        Depth=5    & 1,336 \\
        \bottomrule
    \end{tabular}
    \caption{Number of examples in each of the test datasets for the chaining experiment.}
    \label{tab:chain_dataset_sizes}
\end{table}

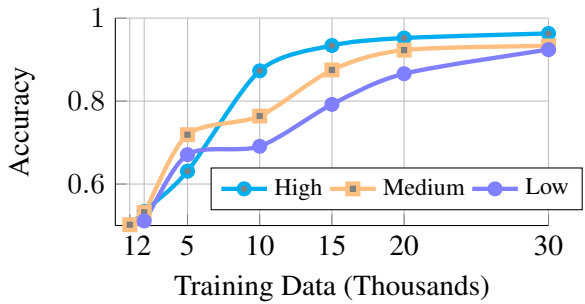
\begin{figure}[tbh]
\hspace{-2ex}
\pgfplotsset{scaled x ticks=true}

\begin{tikzpicture}
\begin{axis}[
  width=5.7cm,
  height=2.75cm,
  scale only axis,
  xmin=0, xmax=30,
  xtick={1,2,5,10,15,20,30},
  xticklabels={1,2,5,10,15,20,30},
  xmajorgrids,
    xlabel= Training Data (Thousands),
    ylabel= Accuracy,
  ymin=0.5, ymax=1.0,
  ymajorgrids,
  axis lines*=left,
  legend style ={ at={(0.22,0.28)},legend columns=5, font = \footnotesize,
        anchor=north west,  
        fill=white,align=left},
    cycle list name=black white,
    smooth
]
\addplot+[color=cyan,line width=1.64pt]
    coordinates{
    (1,0.499)(2,0.534)(5,0.631)(10,0.873) (15,0.934) (20,0.952) (30,0.963)};
    \addlegendentry{High};

    \addplot+[color=orange!50,line width=1.64pt]
    coordinates{
    (1,0.502)(2,0.532)(5,0.719)(10,0.764) (15,0.875) (20,0.923) (30,0.933)};
    \addlegendentry{Medium};
    
    \addplot+[color=blue!50,line width=1.64pt]
    coordinates{
    (1,0.499)(2,0.511)(5,0.671)(10,0.691) (15,0.792) (20,0.866) (30,0.924)};
    \addlegendentry{Low};
    
  \end{axis}
\end{tikzpicture}%
 \caption{Impact of the training data size.}
 \label{fig:training}
\end{figure}

\section{Ablation} 
\label{sec:ablation_exps}

\subsection{Impact of the Data Size}

\paragraph{Setting.} We report the impact of the size of the fine-tuning data on the model performance. As shown in Table~\ref{tab:singles}, the accuracy of the fine-tuned model is higher for rules with higher confidence. We therefore divide the rules in three categories: \textit{High} contains rules with confidence greater than 0.8, \textit{Medium} has rules with confidence between 0.4 and 0.8, and \textit{Low} is for the rest. There are six rules in the \textit{Medium} category and the other two categories have five rules each. For each rule, we fine-tune seven models with 1k, 2k, 5k, 10k, 15k, 20k, and 30k examples.

\paragraph{Results.} Figure~\ref{fig:training} shows that having more training data improves the accuracy in all scenarios. For all categories, there is a sizable increase going from 10k to 15k examples; the impact is smaller for higher values. The highest increase is for rules with high confidence, and rules with medium confidence demonstrate larger increase than low confidence.

\subsection{Role of the Example Format} 

\paragraph{Setting.}
When we teach rules to PLMs, we rely on examples with real names from a fixed pool. However, our goal is to teach PLMs the semantics of the soft rule, not the facts in our examples. Thus, we further design an experiment to assess the impact of the format used in the example facts on the behavior of the model. We distinguish two formats for the generated facts: (\emph{i})~real names such as \textit{Alice} and \textit{IBM}, and (\emph{ii})~letters such as \textit{A} and \textit{B}. We first use each format in fine-tuning and we then test both formats. We end up with two test/train scenarios: one with the same format and one with different formats. For this study, we use just one rule: \textit{child(a,c)} $\wedge$ \textit{parent(c,b)} $\rightarrow$ \textit{spouse(a,b)}, with 30K examples for fine-tuning, and 2k for testing. 

\begin{table}[tbh]
\small
\centering
\begin{tabular}{lcc}
\toprule
 & \bf Train Letter & \bf Train Name \\
\midrule
Test Letter  & .981  & .932        \\
Test Name  &    .977  & .985 \\
\bottomrule
\end{tabular}
\caption{\label{tab:variable}Impact of the example format on accuracy.}
\end{table}

\paragraph{Results.} The results in Table~\ref{tab:variable} show that the model performance does not depend heavily on using the same fact format for training and testing. With examples using letters in training, the results are slightly better in the case with two formats. We ultimately use names for testing and training in our default configuration as it yields better results.

\section{Impact of the Random Seed}
\label{sec:random_seed}

Pre-trained transformers often suffer from instability of the results across multiple reruns with different random seeds. This usually happens with small training datasets~\cite{dodge2020finetuning,mosbach2021on}. In such cases, typically multiple reruns are performed, and the average value over these reruns is reported.

However, the numbers for the main experiments we report in this paper are not averaged over multiple reruns as our datasets are considerably large and the models did not suffer from instability due to random seeds. For example, when we reran \system on a single-rule experiment three times, we obtained accuracy of 0.98959, 0.99551, 0.99636 with a standard deviation of only 0.003. 

Yet, for the small dataset \emph{bAbI}, we observed a much higher standard deviation of 0.17. Thus, in this case we report results that are averaged over ten reruns.

\section{Data Generation Example}
\label{sec:data_gen_ex}

We show an example of data generation for Algorithm~\ref{alg:dg_alg}. For simplicity, here we show an example of a hard rule, i.e.,~one whose confidence is implicitly set to one.\footnote{We show an example of a soft rule in Section~\ref{sec:rule_overlap_example} below.} We begin by setting the values of the input parameters: 

\begin{tcolorbox}[size=small]
\textbf{Algorithm 1 Input:}
\begin{itemize}[noitemsep]
    \item $r=$ \ttfamily{\small{child(A,C) $\wedge$ parent(C,B) $\rightarrow$ spouse(A,B)}}
    \item $n = 8$
    \item $m=5$
    \item $pools=$\{\small{Alice,Bob,Carl,David,Eve}\}
\end{itemize}
\end{tcolorbox}

We set $n = 8$ to generate all the eight hypotheses. We start by generating a set of facts F (line 3), having predicates from the body of the rule with random polarity. We ensure that there are facts that trigger the rule. Their number should not exceed $m$. Here is an example of generated facts $F$:

\begin{tcolorbox}[size=small]
\textbf{Generated Facts F:}
\begin{itemize}[noitemsep]
    \item \textbf{$f_1$:}   \hspace{0.9em}
 \ttfamily{\small{negparent(Eve,Carl)}}
    \item \textbf{$f_2$:} \ttfamily{\small{child(Eve,David)}}
    \item \textbf{$f_3$:} \ttfamily{\small{parent(Carl,Bob)}}
    \item \textbf{$f_4$:} \ttfamily{\small{child(Alice,Carl)}}
\end{itemize}
\end{tcolorbox}

Four facts are generated in total. Facts $f_{3}$ and $f_{4}$ trigger rule $r$. We then feed the rule $r$ and facts $F$ into the \lpmln \ reasoner (line 4). The output $O$ is then:

\begin{tcolorbox}[size=small]
\textbf{\lpmln \ Reasoner Output O:}
\begin{itemize}[noitemsep]
    \item \textbf{$o_1$:} \hspace{0.9em} \ttfamily{\small{child(Eve,David)}}
    \item \textbf{$o_2$:} \ttfamily{\small{child(Alice,Carl)}}
    \item \textbf{$o_3$:} \ttfamily{\small{parent(Carl,Bob)}}
    \item \textbf{$o_4$:} \ttfamily{\small{spouse(Alice,Bob)}}
    \item \textbf{$o_5$:} \ttfamily{\small{negchild(Eve,Carl)}}
\end{itemize}
\end{tcolorbox}

We start generating the hypotheses:

\begin{tcolorbox}[size=small]
\textbf{Generated Hypotheses H:}
\begin{itemize}[noitemsep]
    \item \textbf{$h_1$:} \hspace{0.9em} \ttfamily{\small{child(Eve,David)}}
    \item \textbf{$h_2$:} \ttfamily{\small{child(David,Eve)}}
    
    \item \textbf{$h_3$:} \ttfamily{\small{spouse(Alice,Bob)}}
    \item \textbf{$h_4$:} \ttfamily{\small{negspouse(Alice,Bob)}}
    
    \item \textbf{$h_5$:} \ttfamily{\small{child(David,Carl)}}
    \item \textbf{$h_6$:} \ttfamily{\small{negchild(David,Carl)}}
    
    \item \textbf{$h_7$:} \ttfamily{\small{spouse(Bob,Eve)}}
    \item \textbf{$h_8$:} \ttfamily{\small{negspouse(Bob,Eve)}}
\end{itemize}
\end{tcolorbox}

Hypothesis $h_{1}$ is obtained by sampling from $F$ (line 5), and thus it is a valid hypothesis. Then, the hypothesis $h_{2}$ is generated by altering $h_{1}$ with the function $Alter$ (line 19-22). In this example, since $child$ is not symmetric, $h_{2}$ is produced using a switch of the subject and the object of $h_{1}$ to generate a false hypothesis (line 6).

Hypothesis $h_{3}$ is the outcome of rule $r$ being triggered by facts $f_{3}$ and $f_{4}$ (line 7). In a similar fashion to $h_{2}$, we produce $h_{4}$ (line 8). 

Hypothesis $h_{5}$ is sampled from the universe of all unsatisfied positive facts having a different predicate than that of the rule body (line 9), which makes it an invalid hypothesis, as it is not found in the $O$. Hypothesis $h_{6}$ is the negation of $h_{5}$, and, following CWA, it is a valid hypothesis (line 10).

Finally, hypothesis $h_{7}$ is sampled from the universe of unsatisfied rule-head atoms (line 11), and it is negated to produce hypothesis $h_{8}$.

Overall, we obtain eight different examples represented in symbolic knowledge, where each example contains the set of generated facts $F$, the rule $r$, and a single hypothesis $h_{i}$. The following is one example in symbolic knowledge:

\begin{tcolorbox}[size=small]
\textbf{Example \#1 (Symbolic):}
\begin{itemize}
    \item $Rule \: r=$ \ttfamily{\small{child(A,C) $\wedge$ parent(C,B) $\rightarrow$ spouse(A,B)}}
    \item $Facts \: F:$
          \begin{itemize}[noitemsep]
            \item \textbf{$f_1$:} \ttfamily{\small{negparent(Eve,Carl)}}
            \item \textbf{$f_2$:} \ttfamily{\small{child(Eve,David)}}
            \item \textbf{$f_3$:} \ttfamily{\small{parent(Carl,Bob)}}
            \item \textbf{$f_4$:} \ttfamily{\small{child(Alice,Carl)}}
          \end{itemize}
    \item $Hypothesis \: h_{3}:$ \ttfamily{\small{spouse(Alice,Bob)}}
\end{itemize}
\end{tcolorbox}

We then convert each example to synthetic English using a set of pre-defined templates for the facts and for the rules. Here is the above Example \#1, but now rewritten in synthetic English:

\begin{tcolorbox}[size=small]
\textbf{Example \#1 (Synthetic English):}
\begin{itemize}
    \item $Rule \: r=$ If the child of the first person is the third person, and the parent of the third person is the second person, then the first person is the spouse of the second person.
    \item $Facts \: F:$
          \begin{itemize}
            \item \textbf{$f_1$:} The parent of Eve is not Carl.
            \item \textbf{$f_2$:} The child of Eve is David.
            \item \textbf{$f_3$:} The parent of Carl is Bob.
            \item \textbf{$f_4$:} The child of Alice is Carl.
          \end{itemize}
    \item $Hypothesis \: h_{3}:$ The spouse of Alice is Bob.
\end{itemize}
\end{tcolorbox}

\vspace{6pt}
The $Context$ is defined as the combined set of facts and rule(s). Both $Context$ and $Hypothesis$ are fed as an input to the model.

\begin{tcolorbox}[size=small]
\textbf{Example \#1 (Model Input):}
\begin{itemize}
    \item $Context:$ The parent of Eve is not Carl. The child of Eve is David. If the child of the first person is the third person, and the parent of the third person is the second person, then the first person is the spouse of the second person. The parent of Carl is Bob. The child of Alice is Carl.

    \item $Hypothesis:$ The spouse of Alice is Bob.
\end{itemize}
\end{tcolorbox}

\section{Rule Overlap Example}
\label{sec:rule_overlap_example}

After generating the data for every rule in Figure~\ref{fig:union_exp_rules}, we generate additional examples using combinations of rules. Below, we show how to handle the interaction of two rules: $r_2$ and $r_3$. We follow the procedure in Algorithm~\ref{alg:dg_alg} by generating facts that trigger the rules, but we only take into consideration hypotheses that deal with rule conclusions. 

For example, consider the following facts:

\begin{tcolorbox}[size=small]
\textbf{Generated Facts}
\begin{itemize}[noitemsep]
    \item \textbf{$f_1$:} \hspace{0.9em} \ttfamily{\small{negparent(Eve,Carl)}}
    \item \textbf{$f_2$:} \ttfamily{\small{child(Eve,David)}}
    \item \textbf{$f_3$:} \ttfamily{\small{relative(Eve,David)}}
    \item \textbf{$f_4$:} \ttfamily{\small{predecessor(Eve,David)}}
\end{itemize}
\end{tcolorbox}

We can generate an example that triggers two rules: $f_2$ triggers $r_2$, and $f_{3}$ triggers $r_3$. Feeding the above facts and rules $r_2$ and $r_3$ to the reasoner, we obtain the following output (the numbers in parentheses indicate the likelihood of the triple):

\begin{tcolorbox}[size=small]
\textbf{\lpmln \ Reasoner Output O:}
\begin{itemize}[noitemsep]
    \item \textbf{$o_1$:} \hspace{0.9em} \ttfamily{\small{relative(Eve,David) (1.0)}} 
    \item \textbf{$o_2$:} \ttfamily{\small{child(Eve,David) (1.0)}} 
    \item \textbf{$o_3$:} \ttfamily{\small{negparent(Eve,Carl) (1.0)}}
    \item \textbf{$o_4$:} \ttfamily{\small{spouse(Eve,David) (0.134)}}
    \item \textbf{$o_5$:} \ttfamily{\small{negspouse(Eve,David) (0.55)}} 
    \item \textbf{$o_6$:} \ttfamily{\small{predecessor(Eve,David) (1.0)}} 
\end{itemize}
\end{tcolorbox}

\vspace{6pt}
We produce hypotheses that trigger both rules together. For example, here we generate two hypotheses coming from $o_4$ and $o_5$. The confidence (weight) of a hypothesis is given by the \lpmln \ reasoner. Taking $o_5$ as a hypothesis, we feed the following example to the model:

\begin{tcolorbox}[size=small]
\textbf{Example \#2 (Model Input):}
\begin{itemize}[noitemsep]
    \item $Context:$ The parent of Eve is not Carl. The child of Eve is David. If the child of the first person is the second person, then the first person is not the spouse of the second person. The relative of Eve is David. If the relative of the first person is the second person, then the first person is the spouse of the second person. The predecessor of Eve is David.

    \item $Hypothesis:$ The spouse of Eve is not David.
    
    \item $Weight:$ 0.55
\end{itemize}
\end{tcolorbox}

\vspace{6pt}
We also generate an example, where three rules are triggered: In addition to $r_2$ and $r_3$, $r_5$ is triggered by $f_4$. We then repeat the same procedure to generate the following example:

\begin{tcolorbox}[size=small]
\textbf{Example \#3 (Model Input):}
\begin{itemize}[noitemsep]
    \item $Context:$ The parent of Eve is not Carl. The child of Eve is David. If the child of the first person is the second person, then the first person is not the spouse of the second person. The relative of Eve is David. If the relative of the first person is the second person, then the first person is the spouse of the second person. The predecessor of Eve is David. If the predecessor of the first person is the second person, then the first person is not the spouse of the second person.

    \item $Hypothesis:$ The spouse of Eve is not David.
    
    \item $Weight:$ 0.6
\end{itemize}
\end{tcolorbox}

This procedure is repeated for all combinations of two or more rules. In case when all rules have the same head polarity, we generate a false example by altering the hypothesis and finding the complement of the initial (1-$Weight$) weight.
For example, $r_2$ and $r_5$ can occur together and both have the same rule head, and thus no conflict occurs. The generated valid example would be as follows:

\begin{tcolorbox}[size=small]
\textbf{Example \#4 (Model Input):}
\begin{itemize}
    \item $Context:$ The parent of Eve is not Carl. The child of Eve is David. If the child of the first person is the second person, then the first person is not the spouse of the second person. The relative of Eve is David. The predecessor of Eve is David. If the predecessor of the first person is the second person, then the first person is not the spouse of the second person.

    \item $Hypothesis:$ The spouse of Eve is not David.
    
    \item $Weight:$ 0.64
\end{itemize}
\end{tcolorbox}

An invalid example is generated from the valid example by altering the hypothesis. Here is an invalid example:

\begin{tcolorbox}[size=small]
\textbf{Example \#4 (Model Input):}
\begin{itemize}[noitemsep]
    \item $Context:$ The parent of Eve is not Carl. The child of Eve is David. If the child of the first person is the second person, then the first person is not the spouse of the second person. The relative of Eve is David. The predecessor of Eve is David. If the predecessor of the first person is the second person, then the first person is not the spouse of the second person.

    \item $Hypothesis:$ The spouse of Eve is \st{not} David.
    
    \item $Weight:$ 0.36 (1-0.64)
\end{itemize}
\end{tcolorbox}

\newpage
\section{Rule Chaining Example}
\label{sec:chain_example}
Here is an example that illustrates rule chaining:

\begin{tcolorbox}[size=small]
\textbf{Example \#5 (Symbolic):}
\begin{itemize}[noitemsep]
    \item $Rules \: R=$ 
        \begin{itemize}[noitemsep]
            \item \textbf{$r_1$:} \hspace{0.9em} \ttfamily{\small{child(A,C) $\wedge$ parent(C,B) $\rightarrow$ spouse(A,B)}}
            \item \textbf{$r_2$:} \ttfamily{\small{child(B,A) $\rightarrow$ parent(A,B)}}
        \end{itemize}
    
    \item $Facts \: F:$
          \begin{itemize}[noitemsep]
            \item \textbf{$f_1$:} \hspace{0.9em} \ttfamily{\small{negparent(Eve,Carl)}}
            \item \textbf{$f_2$:} \ttfamily{\small{child(Bob,Carl)}}
            \item \textbf{$f_3$:} \ttfamily{\small{child(Alice,Carl)}}
          \end{itemize}
    \item $Hypothesis \: h:$ \ttfamily{\small{spouse(Alice,Bob)}}
\end{itemize}
\end{tcolorbox}
$f_{2}$ triggers $r_{2}$ which produces $t=$
\begin{flushleft}
    \ttfamily{child(Bob,Carl)}.
\end{flushleft}

$t$ and $f_{3}$ trigger $r_1$ to validate the hypothesis $h$. $r_1$ and $r_2$ have been chained to validate the hypothesis. Since we used two rules to validate the hypothesis, we say that this is a chain of depth $=2$.

\end{document}